\documentclass{article} % For LaTeX2e
\usepackage{iclr2023_conference,times}

\usepackage{amsmath,amsfonts,bm}

\def\eqref#1{equation~\ref{#1}}
\def\1{\bm{1}}

\DeclareMathAlphabet{\mathsfit}{\encodingdefault}{\sfdefault}{m}{sl}
\SetMathAlphabet{\mathsfit}{bold}{\encodingdefault}{\sfdefault}{bx}{n}

\usepackage{hyperref}
\usepackage{url}
\usepackage[utf8]{inputenc} % allow utf-8 input
\usepackage[T1]{fontenc}    % use 8-bit T1 fonts
\usepackage{hyperref}       % hyperlinks
\usepackage{url}            % simple URL typesetting
\usepackage{booktabs}       % professional-quality tables
\usepackage{amsfonts}       % blackboard math symbols
\usepackage{nicefrac}       % compact symbols for 1/2, etc.
\usepackage{microtype}      % microtypography
\usepackage{xcolor}         % colors
\usepackage{algorithm} 
\usepackage{algpseudocode}
\usepackage{graphicx}
\usepackage{amsmath}
\usepackage{subcaption}
\usepackage{comment}
\usepackage{tabularx}

\title{Learning to Communicate and Collaborate in a Competitive Multi-Agent Setup to Clean the Ocean from Macroplastics}

\author{Philipp D. Siedler \\
Aleph Alpha \\
Stuttgart, Germany \\
\texttt{\{p.d.siedler\}@gmail.com} \\
}

\iclrfinalcopy % Uncomment for camera-ready version, but NOT for submission.
\begin{document}

\maketitle

\begin{abstract}
Finding a balance between collaboration and competition is crucial for artificial agents in many real-world applications. We investigate this using a Multi-Agent Reinforcement Learning (MARL) setup on the back of a high-impact problem. The accumulation and yearly growth of plastic in the ocean cause irreparable damage to many aspects of oceanic health and the marina system. To prevent further damage, we need to find ways to reduce macroplastics from known plastic patches in the ocean. Here we propose a Graph Neural Network (GNN) based communication mechanism that increases the agents' observation space. In our custom environment, agents control a plastic collecting vessel. The communication mechanism enables agents to develop a communication protocol using a binary signal. While the goal of the agent collective is to clean up as much as possible, agents are rewarded for the individual amount of macroplastics collected. Hence agents have to learn to communicate effectively while maintaining high individual performance. We compare our proposed communication mechanism with a multi-agent baseline without the ability to communicate. Results show communication enables collaboration and increases collective performance significantly. This means agents have learned the importance of communication and found a balance between collaboration and competition. \textbf{Keywords}: Multi-Agent Reinforcement Learning; Graph Neural Network; Collaboration; Communication; Competition; Proximal Policy Optimization; Ocean Macroplastics;

\end{abstract}

\begin{figure}[ht]
\centering
    \includegraphics[width=\textwidth]{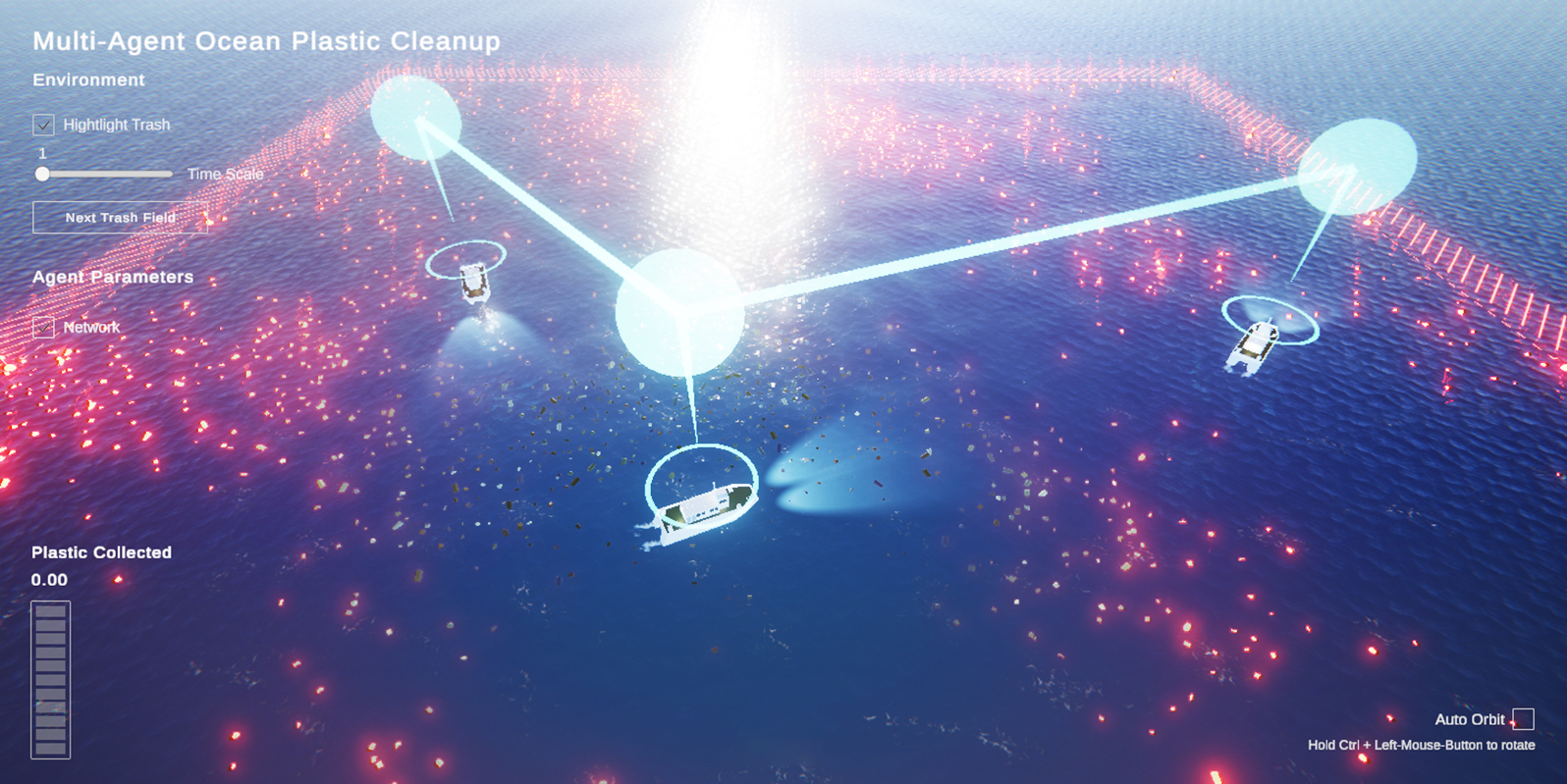}
\caption{Dashboard of multi-agent ocean plastic collector environment. A web application can
be found at: \url{https://philippds-pages.github.io/RL-OPC_WebApp/}.}
\label{fig:thumbnail}
\end{figure}

\newpage

\section{Introduction}

\subsection{Motivation}

% amount of plastic in the ocean
Between 1950 and 2019, plastic production increased to 460 million tonnes yearly \citep{ritchie_plastic_2018}. About 12 million tonnes end up in the ocean \citep{stafford_viewpoint_2019, cressey_bottles_2016, eunomia_plastics_2016}. If the plastic production rate continues to rise, plastic will outweigh fish by 2050 \citep{wef_new_2016}.
% why is this bad for the planet
More than half of the plastic floating in the ocean, i.e. macroplastics, is less dense than water and will not sink to the ground \citep{lebreton_river_2017}. This means that the buoyant plastic mass is within the top meters of the ocean surface \citep{reisser_vertical_2015, kooi_effect_2016}. There are several reasons why this is harmful to marine life, but also to humans and the marine ecosystem. Around 800 coastal but also marine species are affected by plastics in the ocean directly \citep{harding_marine_2016}. 17\% of the species affected are on the International Union for Conservation of Nature (IUCN) list of endangered species \citep{gall_impact_2015}. Animals are specifically affected by a series of causes. Entanglement can restrict movement and the ability to feed, causing infections and may ultimately lead to death \citep{fisheries_entanglement_2021}. Furthermore, animals ingest plastics as they mistake the colour and shape of debris for food. When microplastics get entangled in algae and seaweed, the combination produces an odour that attracts animals \citep{pfaller_odors_2020}. Specifically, microplastics may look like plankton which is food for many species. Even small organisms like polyps living in corals consume microplastics \citep{rotjan_patterns_2019}. Annually 100,000 marine mammals and a million seabirds die because of plastic waste \citep{martin_oceans_2023}. Importantly, when humans consume seafood, they also consume plastic toxins \citep{smith_microplastics_2018}, which can be linked to hormonal abnormalities and development problems \citep{rochester_bisphenol_2013}. Furthermore, there is a concern that plastics in the ocean will degrade to nano-plastics which could enter human cells \citep{mattsson_nano-plastics_2015}. One of the largest accumulations of garbage in the ocean is the Great Pacific Garbage Patch (GPGP), first found and named by Charles J. Moore, a competitive sailor, while returning from a race in 1997. The GPGP measures an area three times the size of France \citep{gruger_trash_2015}, more precisely, the cover surface is estimated at 1.6 million square kilometres \citep{lebreton_evidence_2018}.\\

\subsection{Contribution}

% --> calls for decentralization
To scale the GPGP, a company named The Ocean Cleanup had to utilize a fleet of 30 boats, 652 surface nets and two flights. The size and dynamic properties make this a great use case for a highly distributed system, including multiple agents that have learned to take independent actions. We believe efforts in cleaning up oceans and rivers can benefit from MARL systems. However, for higher planning precision and efficiency, we believe a communication mechanism can contribute significantly. We have been inspired by the world of animals, e.g. the way a dog communicates by wagging its tail or the red colour code of an octopus in an alert state. We propose a highly distributed MARL system with a dynamic GNN communication layer, allowing pairs of agents to observe the garbage density of simulated satellite data and actively communicate signals as part of their action space.

\section{Related Work}

The Ocean Cleanup with their ocean cleaning fleet, but also work on Finite Markov Decision Processes (MDP), e.g. the Recycling Robot by \cite{sutton_reinforcement_2018}, have inspired this work. Furthermore, social aspects like collaboration and communication have been studied in the field of MARL. Communication in Multi-Agent settings is of particular interest as our society and many other distributed systems succeed by collaboration. Work on games, such as Capture The Flag \citep{jaderberg_human-level_2019} and Diplomacy \citep{meta_fundamental_ai_research_diplomacy_team_fair_human-level_2022, kramar_negotiation_2022}, and social dilemmas \citep{foerster_learning_2016, ndousse_emergent_2021, leibo_scalable_2021, agapiou_melting_2022} have investigated social behaviour, e.g. collaboration and competition in multi-agent setups.
We continue work of ours investigating communication, e.g. active sending of information \citep{siedler_power_2021},
active sending and requesting of information in static communication networks \citep{siedler_collaborative_2022}, but also active sending of information from memory in a dynamic communication network \citep{siedler_dynamic_2022}.
Here we investigate active communication in a dynamic but size constraint network and enable the agent collective to develop their own communication protocol. Additional related work can be found in the Appendix \ref{appendix:related-work}. Our learning mechanism design is based on Proximal Policy Optimisation (PPO) \citep{schulman_proximal_2017}, a state-of-the-art, on-policy RL algorithm, that has proven efficient in cooperative MARL settings \citep{yu_surprising_2022}. PPO is defined by two main concepts: 1. Trust region estimation and 2. Advantage estimates. Furthermore, our communication layer is based on a Graph Neural Network (GNN), which operates on graph-structured data, and a message-passing process \citep{gilmer_neural_2017, battaglia_relational_2018}. Additional information on the background of this work can be found in the Appendix \ref{appendix:background}.

\section{Method}

\subsection{Ocean Plastic Collector Environment}

\begin{figure}[ht]
    \begin{subfigure}{0.246\textwidth}
        \centering
        \includegraphics[width=\textwidth]{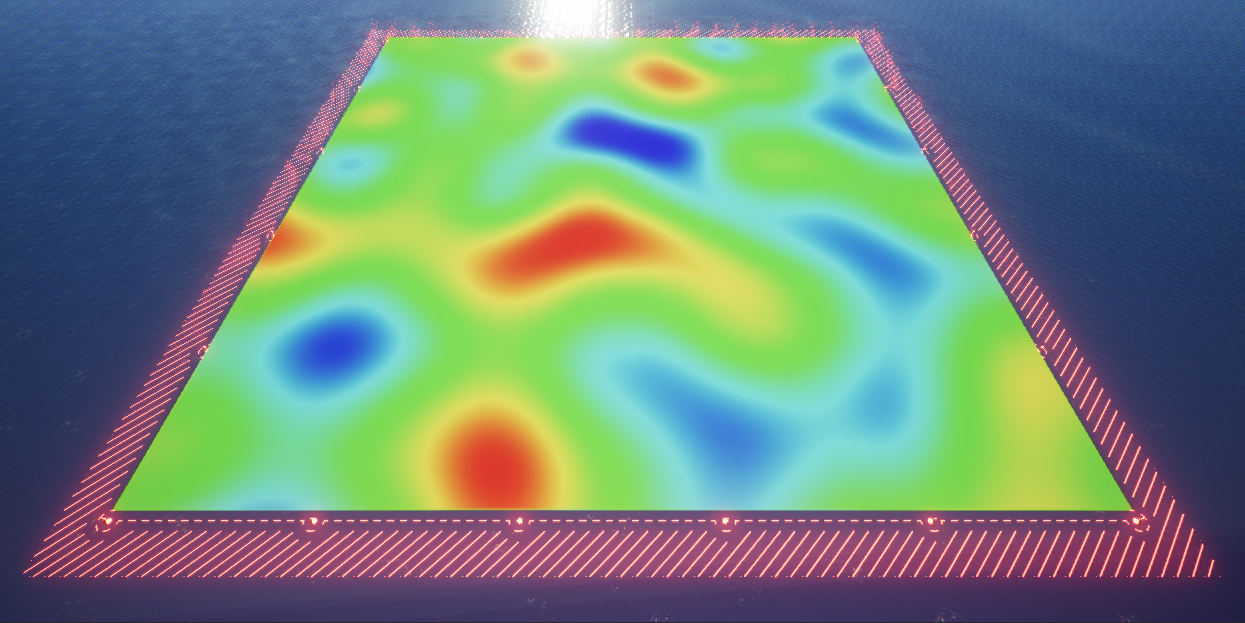}
        \caption{Garbage Heatmap}
        \label{fig:env-heatmap}
        \end{subfigure}
    \begin{subfigure}{0.246\textwidth}
        \centering
        \includegraphics[width=\textwidth]{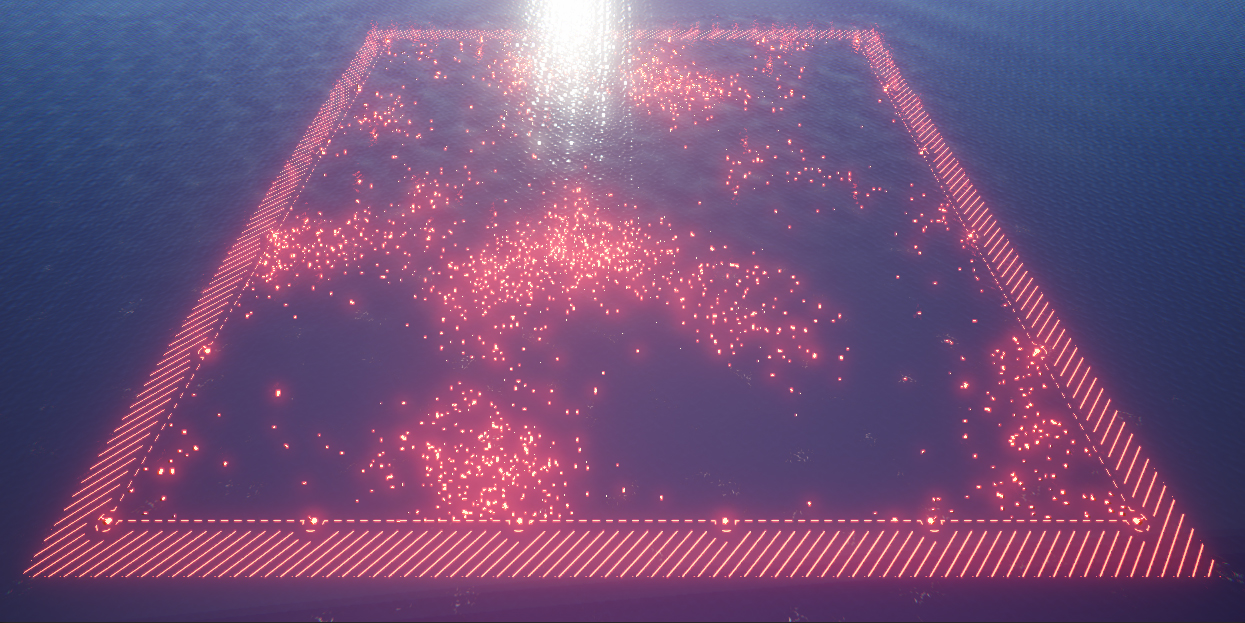}
        \caption{Distributed Garbage}
        \label{fig:env-garbage}
        \end{subfigure}
    \begin{subfigure}{0.246\textwidth}
        \centering
        \includegraphics[width=\textwidth]{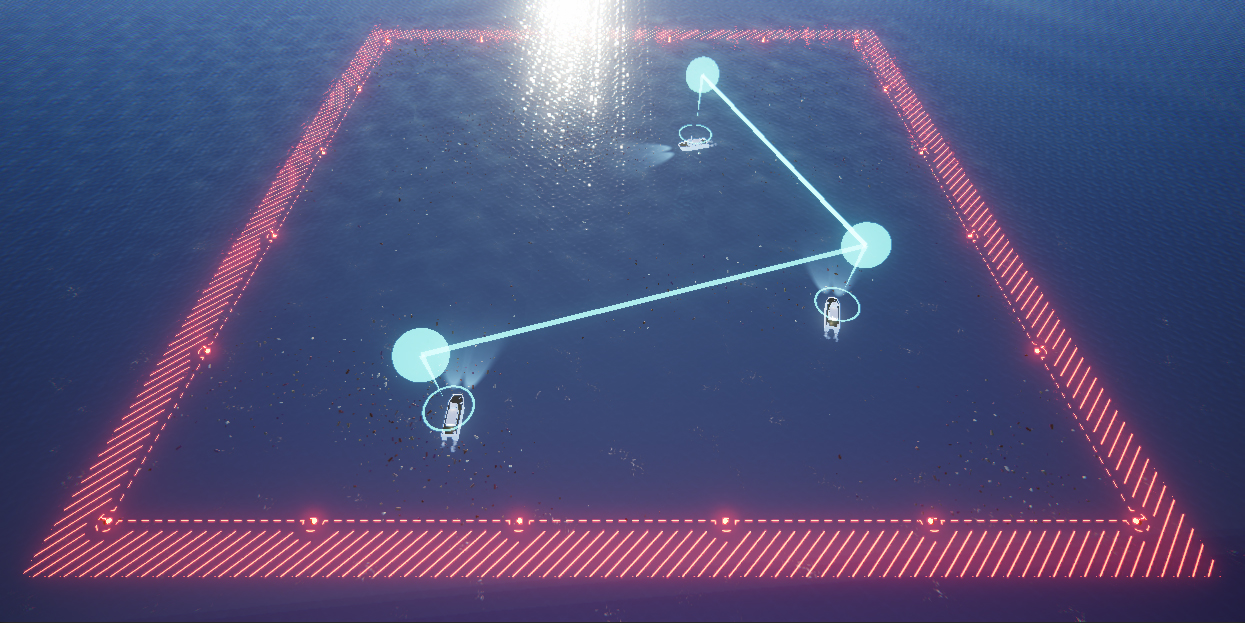}
        \caption{Com. Network}
        \label{fig:env-network}
    \end{subfigure}
    \begin{subfigure}{0.246\textwidth}
        \centering
        \includegraphics[width=\textwidth]{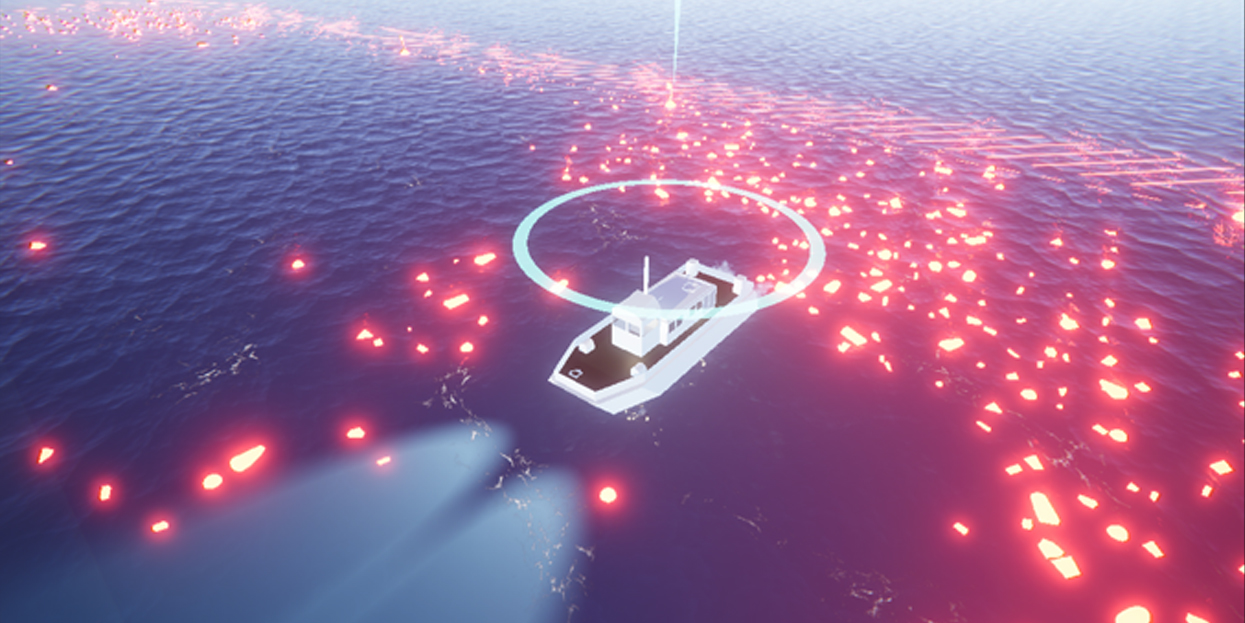}
        \caption{Vessel}
        \label{fig:env-vessel}
    \end{subfigure}
\caption{Ocean Plastic Collector Environment Main Features. Zoomed-in version in the Appendix:\ref{appendix:env-features}}
\label{fig:env-features}
%\vspace{-0.25cm}
\end{figure}

We now describe the features of our custom Ocean Plastic Collector environment built with the Unity game engine. Each environment has multiple training areas which are of size 200m by 200m. A 2D Perlin Noise field \citep{perlin_image_1985} is generated for each scenario and is used to distribute plastic garbage pebbles in the training areas \ref{fig:env-garbage}. Red colour indicates a high and blue low likelihood of pebbles being spawned \ref{fig:env-heatmap}. Each training area includes three agents. A dynamic network connects the three agents if they are equal to or below 100 meters distant from each other \ref{fig:env-network}. Each agent can raise a signal in binary format. The network allows agents to observe the raised signal of neighbouring agents. Lastly, each agent controls an individual vessel and has to navigate to maximize the number of collected floating plastic garbage pebbles \ref{fig:env-vessel}. An episode ends if any vessel crosses the area boundary, crashes into another vessel, or after 5000 steps. At the end of each episode, the noise field and garbage pebbles are reset and a new scenario is generated. The location and rotation of all vessels are randomized.

\subsection{Agent Setup}

\begin{figure}[ht]
\centering
    \includegraphics[width=\textwidth]{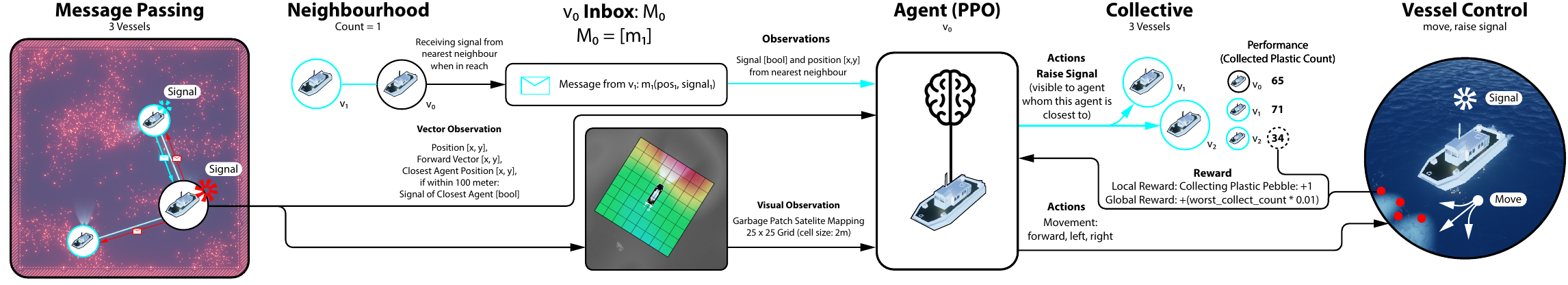}
\caption{MARL Process Diagram. Zoomed-in version in the Appendix \ref{appendix:process}}
\label{fig:process}
\end{figure}

\textbf{Goal:} Each agent must learn to navigate a vessel, collect as many plastic garbage pebbles as possible while staying in the defined area, and avoid crashing into other vessels.

\textbf{Reward Function:} The agent reward function consists of multiple parts. A local positive reward of +1 is given for each plastic garbage pebble collected. Additionally, when a
pebble is collected, a global reward is given. The global reward is calculated as follows: lowest collection count of least performing agent * 0.01. Therefore agents have the incentive to help each other as this collective reward is defined by the least performing agent, i.e. the team is only as strong as its weakest link.

\textbf{Observations:} The observation space consists of vector and visual observations.\\ \textbf{Vector Observations:} The vessel location, simplified as a 2-D vector (x, y) and orientation as a 2-D vector (x, y), the nearest neighbouring agents simplified location as a 2-D vector (x, y) and the nearest neighbours signal [false / true] if within 100 meters, else [false]. The vector observation space size is 14, consisting of 2 stacks of the 7 vector observations.\\
\textbf{Visual Observations:} The visual observation is a grey scale grid of 25 x 25, mimicking a local observation of satellite data mapping of the Greate Pacific Garbage Patch. The observation grid cell size is 2m with a total grid size of 50m x 50m. Each cell holds 0 or 1, existing garbage pebble or no garbage pebble. The visual observation space size is 1250, consisting of 2 stacks of 625 observations. For the visual observation processing, we use a simple encoder that consists of two convolutional layers. Vector and visual observations combined result in a total observation space size of 1264.

\textbf{Actions:} The action space consists of a combination of continuous and discrete actions.\\
\textbf{Continuous Actions:} Each agent has two continuous actions with values in the range of -1 to 1. Continuous actions control the movement of the vessel: Action 0 controls the thrust, and action 1 the rotation, left and right. The movement speed of the vessel is 2, and the rotation speed is 300. The movement speed is translated into force as the vessel is a physics-based object. Torque and directional forces are multiplied by their respective speed factors.\\
\textbf{Discrete Actions:} The discrete action space size is 2 [false / true], allowing the agent to send a signal to the agent it is the nearest neighbour of. There is no qualitative description or pre-defined action plan when receiving a false or true signal, hence the agents as a collective have to develop their own communication protocol.

\section{Experiments}

\begin{table}[h]
%\vspace{-0.5cm}
\caption{Experiment Setup: Multi-Agent without (MA) and with (MAC) the ability to communicate}
\label{experiment-table}
\centering
    \begin{tabularx}{\textwidth}{p{4.2cm} p{1.0cm} p{1.7cm} p{2.2cm} p{3cm}}
    \toprule
    Experiment&Agent(s)&Neighbour(s)&Training Seed&Test Seed\\
    \midrule
    1 Multi-Agent (MA)                  & 3 & 0 & 0-99 (y-shift: 0) & 0-9 (y-shift: 200) \\
    2 MA Communication (MAC)   & 3 & 1 & 0-99 (y-shift: 0) & 0-9 (y-shift: 200) \\
    \bottomrule
    \end{tabularx}
\label{fig:exp}
\end{table}
\begin{figure}[h]
%\vspace{-0.5cm}
\begin{subfigure}{0.246\textwidth}
    \centering
    \includegraphics[width=\textwidth]{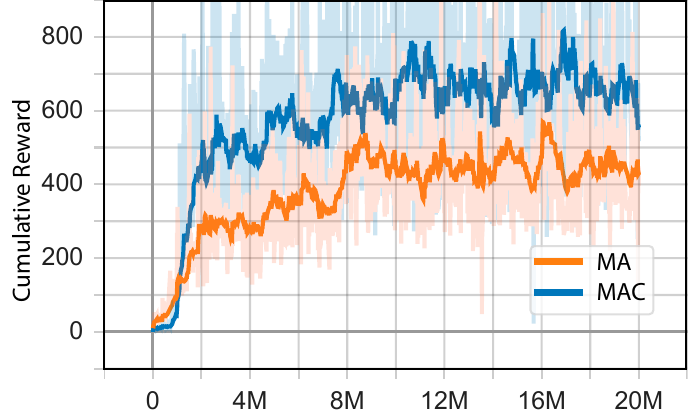}
    \caption{Cml. Reward}
    \label{fig:cml-reward}
    \end{subfigure}
\begin{subfigure}{0.246\textwidth}
    \centering
    \includegraphics[width=\textwidth]{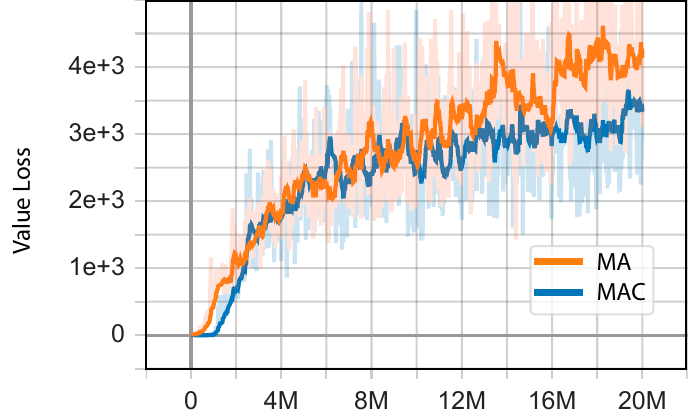}
    \caption{Value Loss}
    \label{fig:value-loss}
    \end{subfigure}
\begin{subfigure}{0.246\textwidth}
    \centering
    \includegraphics[width=\textwidth]{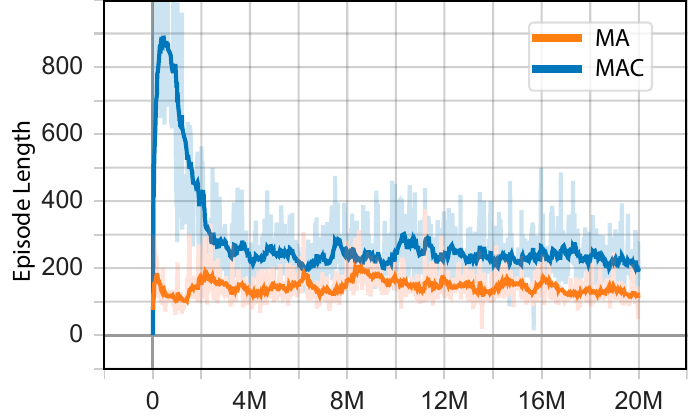}
    \caption{Episode Length}
    \label{fig:episode-length}
\end{subfigure}
\begin{subfigure}{0.246\textwidth}
    \centering
    \includegraphics[width=\textwidth]{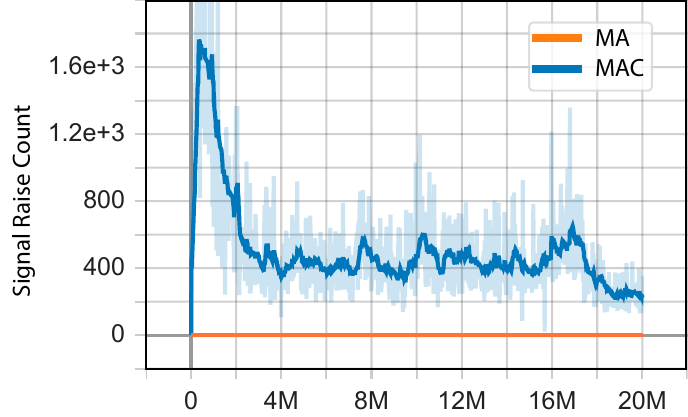}
    \caption{Com. Count}
    \label{fig:signal-raise-count}
\end{subfigure}
\caption{Training graph for 2e7 steps. Zoomed-in data plots in the Appendix: \ref{appendix:train-data-1} and \ref{appendix:train-data-2}}
\label{fig:train}
%\vspace{-0.5cm}
\end{figure}

We performed two experiments. Firstly, a multi-agent setup (MA), without the ability to communicate, serving as the baseline. Secondly, a multi-agent setup with the ability to communicate (MAC) (Table \ref{fig:exp}). In both setups, the observation- and action-space sizes are the same, but the signal for the no-communication setup defaults to false for the two signal-raising actions. The MA and MAC experiments have been trained with three agents on a total 2e7 steps on scenarios with seed 0-99 (Appendix \ref{appendix:env-samples}), and tested on scenarios with seed 0-9 with a noise-map y-shift of 200, ensuring that scenarios in testing did not appear while training. Training data plots can be found in Figure \ref{fig:train}, and training hyperparameters in the Appendix \ref{appendix:hyper}.

\section{Results}

\begin{figure}[ht]
%\vspace{-0.5cm}
    \begin{subfigure}{0.246\textwidth}
        \centering
        \includegraphics[width=\textwidth]{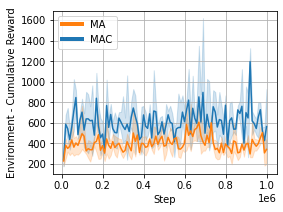}
        \caption{Cml. Reward}
        \label{fig:test-cml-reward}
        \end{subfigure}
    \begin{subfigure}{0.246\textwidth}
        \centering
        \includegraphics[width=\textwidth]{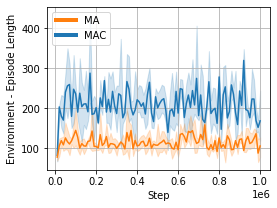}
        \caption{Episode Length}
        \label{fig:test-episode-length}
        \end{subfigure}
    \begin{subfigure}{0.246\textwidth}
        \centering
        \includegraphics[width=\textwidth]{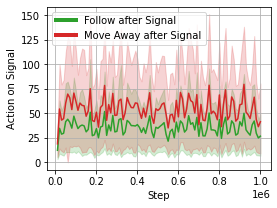}
        \caption{Action on Signal: 1}
        \label{fig:test-action-on-signal}
    \end{subfigure}
    \begin{subfigure}{0.246\textwidth}
        \centering
        \includegraphics[width=\textwidth]{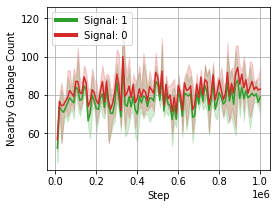}
        \caption{Garbage cnt. Signaling}
        \label{fig:test-garbage-cnt}
    \end{subfigure}
\caption{Testing graphs for 1e6 steps. Zoomed-in version in the Appendix: \ref{appendix:test-data-basics} and \ref{appendix:test-data-signal}}
%\vspace{-0.5cm}
\end{figure}

The results of our experiments (Table \ref{table:results-table}, Figure \ref{fig:cml-reward}) show how communication helps agents increase the cumulative reward by a significant 34\% (Figure \ref{fig:test-cml-reward}). We can further observe that agents have developed a communication protocol where signal 1 (true) is used to communicate to other agents to move away from the signalling agent and signal 0 (false) to communicate to other agents to follow the signalling agent. We can validate this by further investigating the actions taken when a signal is observed. Table \ref{table:com-results-table} and Figure \ref{fig:test-garbage-cnt} show when signal 1 is raised, on average more actions are taken to move away from the agent that raised the signal. Subsequently, when signal 0 is raised, more neighbouring agents take actions to follow the agent that raised the signal (Table \ref{table:com-results-table}, Figure \ref{fig:test-action-on-signal}). Investigating the nearby garbage count when signals are raised, we can find a lower garbage count when the move away signal (1) is raised and a higher garbage count when the follow signal (0) is raised. This validates that the agent collective developed a useful communication protocol. The results on the local and global rewards show how agents are incentivised to help each other and make sure that the least performing agent is improving. Communication increases the local reward by 24\% and the global reward by 41\%. The local reward reflects how many plastic garbage pebbles have been collected. The global reward is the factored performance of the lowest-performing agent in an agent collective. The strong increase in the global reward indicates that agents learned to adapt their behaviour to prioritize the improvement of collective over individual performance.

\begin{table}[h]
\caption{Experiment results. 10 test runs for 1e6 Steps. (↑ better)}
\label{table:results-table}
\centering
    \begin{tabularx}{\textwidth}{p{1.7cm} p{2.9cm} p{2.5cm} p{2.5cm} p{2.5cm}}
    \toprule
    Experiment & Cml. Reward  & Episode Length & Local Reward & Global Reward\\
    \midrule
    1 MA     & 400.67 (±108.27)              & 114.43 (±23.63)           & 137.98 (±30.48)   & 147.53 (±56.44) \\
    2 MAC    & \textbf{606.27 (±205.37.14)}  & \textbf{208.80 (±56.33)}   & \textbf{181.96 (±48.75)}  & \textbf{250.28 (±117.85)} \\
    \bottomrule
    \end{tabularx}
\end{table}

\begin{table}[H]
%\vspace{-0.5cm}
\caption{Communication Metrics of MAC setup}
\label{table:com-results-table}
\centering
\begin{tabularx}{\textwidth}{p{2.3cm} p{2.3cm} p{2.8cm} p{2.3cm} p{2.3cm}}
\toprule
Signal Count & Followed when & Moved Away when & Nearby Garbage Count when & Nearby Garbage Count when\\
per Episode & Signal: 1 & Signal: 1 & Signal: 1 & Signal: 0\\
\midrule
254.77 (±82.57) & 70.60 (±23.21) & 108.48 (±35.07) & 77.52 (±10.13) & 81.75 (±10.29)\\
\bottomrule
\end{tabularx}
\end{table}

\section{Conclusion}

% intro
We have investigated social behaviour in a Multi-Agent setup, on a high-impact problem: communication-enabled collaboration for macroplastic collection in the ocean. Our reward function incentivises agents to develop a useful communication protocol and help the least performing agent in the agent collective. This is leading to a higher number of collected plastic by a large margin.
% environment size VS GPGP
Our environment size is constrained to 200m x 200m, but the GPGP is 1.6 mil. sqm. To bring our approach to the real world, we could increase our training area to fixed subdivisions of the GPGP.
% utilizing existing ship communication capabilities
The development and implementation of autonomous ships are still in their early stages. However, existing ships already have communication systems which could be utilized by the communication mechanism we propose to bring this approach closer to a real-world implementation.
% synthetic data vs real world data
The simulated training environment to distribute garbage and generate an open-ended variety of scenarios was inspired by satellite data. Incorporating real-world data could help improve our simulation's realism and accuracy. However, the use of synthetic data can still be valuable for training, as it allows for a wider range of scenarios to be generated and tested, leading to robustness and generalizability.
% develop communication protocol
We have demonstrated that agents can develop and learn how to use a communication protocol effectively, leading to a high increase in cumulative rewards.
% help the weakest
Furthermore, we have found that rewarding to help the weakest link in an agent collective can lead to agents that care more about the collective than themselves but still achieve high individual performance.
\newpage

\section*{Acknowledgments}
We would like to express our sincere gratitude to Bharathan Balaji for his mentorship, guidance, and valuable insights, which have been instrumental in shaping this work. Heartfelt thanks to Jasmin Arensmeier for her consistent support and patience, as this would not have been possible without her. We also want to thank the workshop organizers and reviewing committee for their efforts and critic, which was very helpful to push the work one step further. More information, video material and an interactive web-app can be found at: \url{https://ai.philippsiedler.com/iclr2023-cooperativeai-gnn-marl-autonomous-ocean-plastic-cleanup/}.

\bibliography{iclr2023_conference}

\begin{thebibliography}{46}
\providecommand{\natexlab}[1]{#1}
\providecommand{\url}[1]{\texttt{#1}}
\expandafter\ifx\csname urlstyle\endcsname\relax
  \providecommand{\doi}[1]{doi: #1}\else
  \providecommand{\doi}{doi: \begingroup \urlstyle{rm}\Url}\fi

\bibitem[Agapiou et~al.(2022)Agapiou, Vezhnevets, Duéñez-Guzmán, Matyas,
  Mao, Sunehag, Köster, Madhushani, Kopparapu, Comanescu, Strouse, Johanson,
  Singh, Haas, Mordatch, Mobbs, and Leibo]{agapiou_melting_2022}
John~P. Agapiou, Alexander~Sasha Vezhnevets, Edgar~A. Duéñez-Guzmán, Jayd
  Matyas, Yiran Mao, Peter Sunehag, Raphael Köster, Udari Madhushani, Kavya
  Kopparapu, Ramona Comanescu, D.~J. Strouse, Michael~B. Johanson, Sukhdeep
  Singh, Julia Haas, Igor Mordatch, Dean Mobbs, and Joel~Z. Leibo.
\newblock Melting {Pot} 2.0, December 2022.
\newblock URL \url{http://arxiv.org/abs/2211.13746}.
\newblock arXiv:2211.13746 [cs].

\bibitem[Baker et~al.(2020)Baker, Kanitscheider, Markov, Wu, Powell, McGrew,
  and Mordatch]{baker_emergent_2020}
Bowen Baker, Ingmar Kanitscheider, Todor Markov, Yi~Wu, Glenn Powell, Bob
  McGrew, and Igor Mordatch.
\newblock Emergent {Tool} {Use} {From} {Multi}-{Agent} {Autocurricula}.
\newblock \emph{arXiv:1909.07528 [cs, stat]}, February 2020.
\newblock URL \url{http://arxiv.org/abs/1909.07528}.
\newblock arXiv: 1909.07528.

\bibitem[Battaglia et~al.(2018)Battaglia, Hamrick, Bapst, Sanchez-Gonzalez,
  Zambaldi, Malinowski, Tacchetti, Raposo, Santoro, Faulkner, Gulcehre, Song,
  Ballard, Gilmer, Dahl, Vaswani, Allen, Nash, Langston, Dyer, Heess, Wierstra,
  Kohli, Botvinick, Vinyals, Li, and Pascanu]{battaglia_relational_2018}
Peter~W. Battaglia, Jessica~B. Hamrick, Victor Bapst, Alvaro Sanchez-Gonzalez,
  Vinicius Zambaldi, Mateusz Malinowski, Andrea Tacchetti, David Raposo, Adam
  Santoro, Ryan Faulkner, Caglar Gulcehre, Francis Song, Andrew Ballard, Justin
  Gilmer, George Dahl, Ashish Vaswani, Kelsey Allen, Charles Nash, Victoria
  Langston, Chris Dyer, Nicolas Heess, Daan Wierstra, Pushmeet Kohli, Matt
  Botvinick, Oriol Vinyals, Yujia Li, and Razvan Pascanu.
\newblock Relational inductive biases, deep learning, and graph networks.
\newblock \emph{arXiv:1806.01261 [cs, stat]}, October 2018.
\newblock URL \url{http://arxiv.org/abs/1806.01261}.
\newblock arXiv: 1806.01261.

\bibitem[Biermann et~al.(2019)Biermann, Martinez~Vincente, Sailley, Mata, and
  Steele]{biermann_towards_2019}
Lauren Biermann, Victor Martinez~Vincente, Sevrine Sailley, Aser Mata, and
  Christopher Steele.
\newblock Towards a method for detecting macroplastics by satellite: examining
  {Sentinel}-2 earth observation data for floating debris in the coastal zone.
\newblock pp.\  17469, 2019.
\newblock URL \url{https://ui.adsabs.harvard.edu/abs/2019EGUGA..2117469B}.
\newblock Conference Name: EGU General Assembly Conference Abstracts ADS
  Bibcode: 2019EGUGA..2117469B.

\bibitem[Clearbot(2022)]{clearbot_this_2022}
Clearbot.
\newblock This {AI}-enabled robotic boat cleans up harbors and rivers to keep
  plastic trash out of the ocean, 2022.
\newblock URL \url{https://news.microsoft.com/apac/features/}.

\bibitem[Cressey(2016)]{cressey_bottles_2016}
Daniel Cressey.
\newblock Bottles, bags, ropes and toothbrushes: the struggle to track ocean
  plastics.
\newblock \emph{Nature}, 536\penalty0 (7616):\penalty0 263--265, August 2016.
\newblock ISSN 1476-4687.
\newblock \doi{10.1038/536263a}.
\newblock URL \url{https://www.nature.com/articles/536263a}.
\newblock Number: 7616 Publisher: Nature Publishing Group.

\bibitem[Eunomia(2016)]{eunomia_plastics_2016}
Eunomia.
\newblock Plastics in the {Marine} {Environment}, 2016.
\newblock URL
  \url{https://www.eunomia.co.uk/reports-tools/plastics-in-the-marine-environment/}.

\bibitem[Fisheries(2021)]{fisheries_entanglement_2021}
NOAA Fisheries.
\newblock Entanglement of {Marine} {Life}: {Risks} and {Response} {\textbar}
  {NOAA} {Fisheries}, January 2021.
\newblock URL
  \url{https://www.fisheries.noaa.gov/insight/entanglement-marine-life-risks-and-response}.
\newblock Archive Location: National.

\bibitem[Foerster et~al.(2016)Foerster, Assael, de~Freitas, and
  Whiteson]{foerster_learning_2016}
Jakob Foerster, Ioannis~Alexandros Assael, Nando de~Freitas, and Shimon
  Whiteson.
\newblock Learning to {Communicate} with {Deep} {Multi}-{Agent} {Reinforcement}
  {Learning}.
\newblock In \emph{Advances in {Neural} {Information} {Processing} {Systems}},
  volume~29, Barcelona, Spain, 2016. Curran Associates, Inc.
\newblock URL
  \url{https://papers.nips.cc/paper/2016/hash/c7635bfd99248a2cdef8249ef7bfbef4-Abstract.html}.

\bibitem[Gall \& Thompson(2015)Gall and Thompson]{gall_impact_2015}
S.~C. Gall and R.~C. Thompson.
\newblock The impact of debris on marine life.
\newblock \emph{Marine Pollution Bulletin}, 92\penalty0 (1):\penalty0 170--179,
  March 2015.
\newblock ISSN 0025-326X.
\newblock \doi{10.1016/j.marpolbul.2014.12.041}.
\newblock URL
  \url{https://www.sciencedirect.com/science/article/pii/S0025326X14008571}.

\bibitem[Gilmer et~al.(2017)Gilmer, Schoenholz, Riley, Vinyals, and
  Dahl]{gilmer_neural_2017}
Justin Gilmer, Samuel~S. Schoenholz, Patrick~F. Riley, Oriol Vinyals, and
  George~E. Dahl.
\newblock Neural {Message} {Passing} for {Quantum} {Chemistry}.
\newblock \emph{arXiv:1704.01212 [cs]}, June 2017.
\newblock URL \url{http://arxiv.org/abs/1704.01212}.
\newblock arXiv: 1704.01212.

\bibitem[Gruger(2015)]{gruger_trash_2015}
Sally Gruger.
\newblock {TRASH} {TALK}: {What} is the {Great} {Pacific} {Garbage} {Patch}?
  {\textbar} {OR}\&{R}'s {Marine} {Debris} {Program}, July 2015.
\newblock URL
  \url{https://marinedebris.noaa.gov/videos/trash-talk-what-great-pacific-garbage-patch-0}.

\bibitem[Harding(2016)]{harding_marine_2016}
Simon Harding.
\newblock Marine {Debris}: {Understanding}, {Preventing} and {Mitigating} the
  {Significant} {Adverse} {Impacts} on {Marine} and {Coastal} {Biodiversity}.
\newblock Report, Secretariat of the Convention on Biological Diversity, 2016.
\newblock URL
  \url{https://repository.oceanbestpractices.org/handle/11329/1625}.
\newblock Accepted: 2021-07-16T19:39:20Z ISBN: 9789292256258 Journal
  Abbreviation: Montreal, Canada.

\bibitem[HYCOM(2023)]{hycom_hycom_2023}
HYCOM.
\newblock {HYCOM}, 2023.
\newblock URL \url{https://www.hycom.org/}.

\bibitem[Jaderberg et~al.(2019)Jaderberg, Czarnecki, Dunning, Marris, Lever,
  Castaneda, Beattie, Rabinowitz, Morcos, Ruderman, Sonnerat, Green, Deason,
  Leibo, Silver, Hassabis, Kavukcuoglu, and
  Graepel]{jaderberg_human-level_2019}
Max Jaderberg, Wojciech~M. Czarnecki, Iain Dunning, Luke Marris, Guy Lever,
  Antonio~Garcia Castaneda, Charles Beattie, Neil~C. Rabinowitz, Ari~S. Morcos,
  Avraham Ruderman, Nicolas Sonnerat, Tim Green, Louise Deason, Joel~Z. Leibo,
  David Silver, Demis Hassabis, Koray Kavukcuoglu, and Thore Graepel.
\newblock Human-level performance in first-person multiplayer games with
  population-based deep reinforcement learning.
\newblock \emph{Science}, 364\penalty0 (6443):\penalty0 859--865, May 2019.
\newblock ISSN 0036-8075, 1095-9203.
\newblock \doi{10.1126/science.aau6249}.
\newblock URL \url{http://arxiv.org/abs/1807.01281}.
\newblock arXiv:1807.01281 [cs, stat].

\bibitem[Juliani et~al.(2020)Juliani, Berges, Teng, Cohen, Harper, Elion, Goy,
  Gao, Henry, Mattar, and Lange]{juliani_unity_2020}
Arthur Juliani, Vincent-Pierre Berges, Ervin Teng, Andrew Cohen, Jonathan
  Harper, Chris Elion, Chris Goy, Yuan Gao, Hunter Henry, Marwan Mattar, and
  Danny Lange.
\newblock Unity: {A} {General} {Platform} for {Intelligent} {Agents}, May 2020.
\newblock URL \url{http://arxiv.org/abs/1809.02627}.
\newblock arXiv:1809.02627 [cs, stat].

\bibitem[Kooi et~al.(2016)Kooi, Reisser, Slat, Ferrari, Schmid, Cunsolo,
  Brambini, Noble, Sirks, Linders, Schoeneich-Argent, and
  Koelmans]{kooi_effect_2016}
Merel Kooi, Julia Reisser, Boyan Slat, Francesco~F. Ferrari, Moritz~S. Schmid,
  Serena Cunsolo, Roberto Brambini, Kimberly Noble, Lys-Anne Sirks, Theo E.~W.
  Linders, Rosanna~I. Schoeneich-Argent, and Albert~A. Koelmans.
\newblock The effect of particle properties on the depth profile of buoyant
  plastics in the ocean.
\newblock \emph{Scientific Reports}, 6\penalty0 (1):\penalty0 33882, October
  2016.
\newblock ISSN 2045-2322.
\newblock \doi{10.1038/srep33882}.
\newblock URL \url{https://www.nature.com/articles/srep33882}.
\newblock Number: 1 Publisher: Nature Publishing Group.

\bibitem[Kramár et~al.(2022)Kramár, Eccles, Gemp, Tacchetti, McKee,
  Malinowski, Graepel, and Bachrach]{kramar_negotiation_2022}
János Kramár, Tom Eccles, Ian Gemp, Andrea Tacchetti, Kevin~R. McKee, Mateusz
  Malinowski, Thore Graepel, and Yoram Bachrach.
\newblock Negotiation and honesty in artificial intelligence methods for the
  board game of {Diplomacy}.
\newblock \emph{Nature Communications}, 13\penalty0 (1):\penalty0 7214,
  December 2022.
\newblock ISSN 2041-1723.
\newblock \doi{10.1038/s41467-022-34473-5}.
\newblock URL \url{https://www.nature.com/articles/s41467-022-34473-5}.
\newblock Number: 1 Publisher: Nature Publishing Group.

\bibitem[Lebreton et~al.(2018)Lebreton, Slat, Ferrari, Sainte-Rose, Aitken,
  Marthouse, Hajbane, Cunsolo, Schwarz, Levivier, Noble, Debeljak, Maral,
  Schoeneich-Argent, Brambini, and Reisser]{lebreton_evidence_2018}
L.~Lebreton, B.~Slat, F.~Ferrari, B.~Sainte-Rose, J.~Aitken, R.~Marthouse,
  S.~Hajbane, S.~Cunsolo, A.~Schwarz, A.~Levivier, K.~Noble, P.~Debeljak,
  H.~Maral, R.~Schoeneich-Argent, R.~Brambini, and J.~Reisser.
\newblock Evidence that the {Great} {Pacific} {Garbage} {Patch} is rapidly
  accumulating plastic.
\newblock \emph{Scientific Reports}, 8\penalty0 (1):\penalty0 4666, March 2018.
\newblock ISSN 2045-2322.
\newblock \doi{10.1038/s41598-018-22939-w}.
\newblock URL \url{https://www.nature.com/articles/s41598-018-22939-w}.
\newblock Number: 1 Publisher: Nature Publishing Group.

\bibitem[Lebreton et~al.(2017)Lebreton, van~der Zwet, Damsteeg, Slat, Andrady,
  and Reisser]{lebreton_river_2017}
Laurent C.~M. Lebreton, Joost van~der Zwet, Jan-Willem Damsteeg, Boyan Slat,
  Anthony Andrady, and Julia Reisser.
\newblock River plastic emissions to the world’s oceans.
\newblock \emph{Nature Communications}, 8\penalty0 (1):\penalty0 15611, June
  2017.
\newblock ISSN 2041-1723.
\newblock \doi{10.1038/ncomms15611}.
\newblock URL \url{https://www.nature.com/articles/ncomms15611}.
\newblock Number: 1 Publisher: Nature Publishing Group.

\bibitem[Leibo et~al.(2021)Leibo, Duéñez-Guzmán, Vezhnevets, Agapiou,
  Sunehag, Koster, Matyas, Beattie, Mordatch, and Graepel]{leibo_scalable_2021}
Joel~Z. Leibo, Edgar Duéñez-Guzmán, Alexander~Sasha Vezhnevets, John~P.
  Agapiou, Peter Sunehag, Raphael Koster, Jayd Matyas, Charles Beattie, Igor
  Mordatch, and Thore Graepel.
\newblock Scalable {Evaluation} of {Multi}-{Agent} {Reinforcement} {Learning}
  with {Melting} {Pot}, July 2021.
\newblock URL \url{http://arxiv.org/abs/2107.06857}.
\newblock arXiv:2107.06857 [cs].

\bibitem[{Martin}(2023)]{martin_oceans_2023}
{Martin}.
\newblock Oceans, 2023.
\newblock URL \url{https://www.un.org/sustainabledevelopment/oceans/}.

\bibitem[Mattsson et~al.(2015)Mattsson, Hansson, and
  Cedervall]{mattsson_nano-plastics_2015}
K.~Mattsson, L.-A. Hansson, and T.~Cedervall.
\newblock Nano-plastics in the aquatic environment.
\newblock \emph{Environmental Science. Processes \& Impacts}, 17\penalty0
  (10):\penalty0 1712--1721, October 2015.
\newblock ISSN 2050-7895.
\newblock \doi{10.1039/c5em00227c}.

\bibitem[{Meta Fundamental AI Research Diplomacy Team (FAIR)} et~al.(2022){Meta
  Fundamental AI Research Diplomacy Team (FAIR)}, Bakhtin, Brown, Dinan,
  Farina, Flaherty, Fried, Goff, Gray, Hu, Jacob, Komeili, Konath, Kwon, Lerer,
  Lewis, Miller, Mitts, Renduchintala, Roller, Rowe, Shi, Spisak, Wei, Wu,
  Zhang, and
  Zijlstra]{meta_fundamental_ai_research_diplomacy_team_fair_human-level_2022}
{Meta Fundamental AI Research Diplomacy Team (FAIR)}, Anton Bakhtin, Noam
  Brown, Emily Dinan, Gabriele Farina, Colin Flaherty, Daniel Fried, Andrew
  Goff, Jonathan Gray, Hengyuan Hu, Athul~Paul Jacob, Mojtaba Komeili, Karthik
  Konath, Minae Kwon, Adam Lerer, Mike Lewis, Alexander~H. Miller, Sasha Mitts,
  Adithya Renduchintala, Stephen Roller, Dirk Rowe, Weiyan Shi, Joe Spisak,
  Alexander Wei, David Wu, Hugh Zhang, and Markus Zijlstra.
\newblock Human-level play in the game of {Diplomacy} by combining language
  models with strategic reasoning.
\newblock \emph{Science}, 378\penalty0 (6624):\penalty0 1067--1074, December
  2022.
\newblock \doi{10.1126/science.ade9097}.
\newblock URL \url{https://www.science.org/doi/10.1126/science.ade9097}.
\newblock Publisher: American Association for the Advancement of Science.

\bibitem[NCEI(2023)]{ncei_global_2023}
NCEI.
\newblock Global {Forecast} {System} ({GFS}) {\textbar} {National} {Centers}
  for {Environmental} {Information} ({NCEI}), 2023.
\newblock URL
  \url{https://www.ncei.noaa.gov/products/weather-climate-models/global-forecast}.

\bibitem[Ndousse et~al.(2021)Ndousse, Eck, Levine, and
  Jaques]{ndousse_emergent_2021}
Kamal~K. Ndousse, Douglas Eck, Sergey Levine, and Natasha Jaques.
\newblock Emergent {Social} {Learning} via {Multi}-agent {Reinforcement}
  {Learning}.
\newblock In \emph{Proceedings of the 38th {International} {Conference} on
  {Machine} {Learning}}, pp.\  7991--8004. PMLR, July 2021.
\newblock URL \url{https://proceedings.mlr.press/v139/ndousse21a.html}.
\newblock ISSN: 2640-3498.

\bibitem[OpenAI(2021)]{openai_proximal_2021}
Spinning~Up OpenAI.
\newblock Proximal {Policy} {Optimization} — {Spinning} {Up} documentation,
  2021.
\newblock URL
  \url{https://spinningup.openai.com/en/latest/algorithms/ppo.html}.

\bibitem[Perlin(1985)]{perlin_image_1985}
Ken Perlin.
\newblock An image synthesizer.
\newblock \emph{ACM SIGGRAPH Computer Graphics}, 19\penalty0 (3):\penalty0
  287--296, July 1985.
\newblock ISSN 0097-8930.
\newblock \doi{10.1145/325165.325247}.
\newblock URL \url{https://doi.org/10.1145/325165.325247}.

\bibitem[Pfaller et~al.(2020)Pfaller, Goforth, Gil, Savoca, and
  Lohmann]{pfaller_odors_2020}
Joseph~B. Pfaller, Kayla~M. Goforth, Michael~A. Gil, Matthew~S. Savoca, and
  Kenneth~J. Lohmann.
\newblock Odors from marine plastic debris elicit foraging behavior in sea
  turtles.
\newblock \emph{Current Biology}, 30\penalty0 (5):\penalty0 R213--R214, March
  2020.
\newblock ISSN 0960-9822.
\newblock \doi{10.1016/j.cub.2020.01.071}.
\newblock URL
  \url{https://www.cell.com/current-biology/abstract/S0960-9822(20)30115-9}.
\newblock Publisher: Elsevier.

\bibitem[Reisser et~al.(2015)Reisser, Slat, Noble, du~Plessis, Epp, Proietti,
  de~Sonneville, Becker, and Pattiaratchi]{reisser_vertical_2015}
J.~Reisser, B.~Slat, K.~Noble, K.~du~Plessis, M.~Epp, M.~Proietti,
  J.~de~Sonneville, T.~Becker, and C.~Pattiaratchi.
\newblock The vertical distribution of buoyant plastics at sea: an
  observational study in the {North} {Atlantic} {Gyre}.
\newblock \emph{Biogeosciences}, 12\penalty0 (4):\penalty0 1249--1256, February
  2015.
\newblock ISSN 1726-4170.
\newblock \doi{10.5194/bg-12-1249-2015}.
\newblock URL \url{https://bg.copernicus.org/articles/12/1249/2015/}.
\newblock Publisher: Copernicus GmbH.

\bibitem[Ritchie \& Roser(2018)Ritchie and Roser]{ritchie_plastic_2018}
Hannah Ritchie and Max Roser.
\newblock Plastic {Pollution}.
\newblock \emph{Our World in Data}, September 2018.
\newblock URL \url{https://ourworldindata.org/plastic-pollution}.

\bibitem[Rochester(2013)]{rochester_bisphenol_2013}
Johanna~R. Rochester.
\newblock Bisphenol {A} and human health: {A} review of the literature.
\newblock \emph{Reproductive Toxicology}, 42:\penalty0 132--155, December 2013.
\newblock ISSN 0890-6238.
\newblock \doi{10.1016/j.reprotox.2013.08.008}.
\newblock URL
  \url{https://www.sciencedirect.com/science/article/pii/S0890623813003456}.

\bibitem[Rotjan et~al.(2019)Rotjan, Sharp, Gauthier, Yelton, Lopez, Carilli,
  Kagan, and Urban-Rich]{rotjan_patterns_2019}
Randi~D. Rotjan, Koty~H. Sharp, Anna~E. Gauthier, Rowan Yelton, Eliya M.~Baron
  Lopez, Jessica Carilli, Jonathan~C. Kagan, and Juanita Urban-Rich.
\newblock Patterns, dynamics and consequences of microplastic ingestion by the
  temperate coral, {Astrangia} poculata.
\newblock \emph{Proceedings of the Royal Society B: Biological Sciences},
  286\penalty0 (1905):\penalty0 20190726, June 2019.
\newblock \doi{10.1098/rspb.2019.0726}.
\newblock URL
  \url{https://royalsocietypublishing.org/doi/10.1098/rspb.2019.0726}.
\newblock Publisher: Royal Society.

\bibitem[Schulman et~al.(2017)Schulman, Wolski, Dhariwal, Radford, and
  Klimov]{schulman_proximal_2017}
John Schulman, Filip Wolski, Prafulla Dhariwal, Alec Radford, and Oleg Klimov.
\newblock Proximal {Policy} {Optimization} {Algorithms}.
\newblock \emph{arXiv:1707.06347 [cs]}, August 2017.
\newblock URL \url{http://arxiv.org/abs/1707.06347}.
\newblock arXiv: 1707.06347.

\bibitem[Siedler(2021)]{siedler_power_2021}
Philipp~Dominic Siedler.
\newblock The {Power} of {Communication} in a {Distributed} {Multi}-{Agent}
  {System}.
\newblock \emph{arXiv:2111.15611 [cs]}, December 2021.
\newblock URL \url{http://arxiv.org/abs/2111.15611}.
\newblock arXiv: 2111.15611.

\bibitem[Siedler(2022{\natexlab{a}})]{siedler_collaborative_2022}
Philipp~Dominic Siedler.
\newblock Collaborative {Auto}-{Curricula} {Multi}-{Agent} {Reinforcement}
  {Learning} with {Graph} {Neural} {Network} {Communication} {Layer} for
  {Open}-ended {Wildfire}-{Management} {Resource} {Distribution}, April
  2022{\natexlab{a}}.
\newblock URL \url{http://arxiv.org/abs/2204.11350}.
\newblock arXiv:2204.11350 [cs].

\bibitem[Siedler(2022{\natexlab{b}})]{siedler_dynamic_2022}
Philipp~Dominic Siedler.
\newblock Dynamic {Collaborative} {Multi}-{Agent} {Reinforcement} {Learning}
  {Communication} for {Autonomous} {Drone} {Reforestation}, November
  2022{\natexlab{b}}.
\newblock URL \url{https://arxiv.org/abs/2211.15414v1}.
\newblock arXiv:2211.15414 version: 1.

\bibitem[Smith et~al.(2018)Smith, Love, Rochman, and
  Neff]{smith_microplastics_2018}
Madeleine Smith, David~C. Love, Chelsea~M. Rochman, and Roni~A. Neff.
\newblock Microplastics in {Seafood} and the {Implications} for {Human}
  {Health}.
\newblock \emph{Current Environmental Health Reports}, 5\penalty0 (3):\penalty0
  375--386, 2018.
\newblock ISSN 2196-5412.
\newblock \doi{10.1007/s40572-018-0206-z}.
\newblock URL \url{https://www.ncbi.nlm.nih.gov/pmc/articles/PMC6132564/}.

\bibitem[Stafford \& Jones(2019)Stafford and Jones]{stafford_viewpoint_2019}
Richard Stafford and Peter J.~S. Jones.
\newblock Viewpoint – {Ocean} plastic pollution: {A} convenient but
  distracting truth?
\newblock \emph{Marine Policy}, 103:\penalty0 187--191, May 2019.
\newblock ISSN 0308-597X.
\newblock \doi{10.1016/j.marpol.2019.02.003}.
\newblock URL
  \url{https://www.sciencedirect.com/science/article/pii/S0308597X1830681X}.

\bibitem[Sutton \& Barto(2018)Sutton and Barto]{sutton_reinforcement_2018}
Richard~S. Sutton and Andrew~G. Barto.
\newblock \emph{Reinforcement {Learning}: {An} {Introduction}}.
\newblock A Bradford Book, Cambridge, MA, USA, 2018.
\newblock ISBN 978-0-262-03924-6.

\bibitem[Torres et~al.(2018)Torres, Klein, Menemenlis, Qiu, Su, Wang, Chen, and
  Fu]{torres_partitioning_2018}
Hector Torres, Patrice Klein, Dimitris Menemenlis, Bo~Qiu, Zhan Su, Jinbo Wang,
  Shuiming Chen, and Lee-Lueng Fu.
\newblock Partitioning {Ocean} {Motions} {Into} {Balanced} {Motions} and
  {Internal} {Gravity} {Waves}: {A} {Modeling} {Study} in {Anticipation} of
  {Future} {Space} {Missions}.
\newblock \emph{Journal of Geophysical Research: Oceans}, 123, October 2018.
\newblock \doi{10.1029/2018JC014438}.

\bibitem[Wang \& Raj(2017)Wang and Raj]{wang_origin_2017}
Haohan Wang and Bhiksha Raj.
\newblock On the {Origin} of {Deep} {Learning}.
\newblock \emph{arXiv:1702.07800 [cs, stat]}, March 2017.
\newblock URL \url{http://arxiv.org/abs/1702.07800}.
\newblock arXiv: 1702.07800.

\bibitem[Wang et~al.(2019)Wang, Hao, Wang, and Taylor]{wang_achieving_2019}
Weixun Wang, Jianye Hao, Yixi Wang, and Matthew Taylor.
\newblock Achieving cooperation through deep multiagent reinforcement learning
  in sequential prisoner's dilemmas.
\newblock In \emph{Proceedings of the {First} {International} {Conference} on
  {Distributed} {Artificial} {Intelligence}}, {DAI} '19, pp.\  1--7, New York,
  NY, USA, October 2019. Association for Computing Machinery.
\newblock ISBN 978-1-4503-7656-3.
\newblock \doi{10.1145/3356464.3357712}.
\newblock URL \url{https://doi.org/10.1145/3356464.3357712}.

\bibitem[WEF(2016)]{wef_new_2016}
WEF.
\newblock The {New} {Plastics} {Economy}: {Rethinking} the future of plastics,
  2016.
\newblock URL
  \url{https://www.weforum.org/reports/the-new-plastics-economy-rethinking-the-future-of-plastics/}.

\bibitem[Yu et~al.(2022)Yu, Velu, Vinitsky, Gao, Wang, Bayen, and
  Wu]{yu_surprising_2022}
Chao Yu, Akash Velu, Eugene Vinitsky, Jiaxuan Gao, Yu~Wang, Alexandre Bayen,
  and Yi~Wu.
\newblock The {Surprising} {Effectiveness} of {PPO} in {Cooperative}
  {Multi}-{Agent} {Games}.
\newblock 2022.

\bibitem[Zychlinski(2019)]{zychlinski_complete_2019}
Shaked Zychlinski.
\newblock The {Complete} {Reinforcement} {Learning} {Dictionary}, November
  2019.
\newblock URL
  \url{https://towardsdatascience.com/the-complete-reinforcement-learning-dictionary-e16230b7d24e}.

\end{thebibliography}
\bibliographystyle{iclr2023_conference}

\newpage

\appendix
\section{Appendix}

\section{Related Work}
\label{appendix:related-work}

\textbf{Autonomous Garbage Collection:} Some of the most groundbreaking ideas and milestones in Reinforcement Learning (RL) are the basis for MARL \citep{wang_origin_2017}. Therefore, inspiration but also a common example of Finite Markov Decision Processes (MDP) is the Recycling Robot, as part of the Introduction to Reinforcement Learning book by \cite{sutton_reinforcement_2018}. Closer to our use case, however, is the Unity ML-Agents Food Collector environment \citep{juliani_unity_2020}. Examples of autonomous vehicles collecting garbage can also be found in industry and academia, such as the Clearbot Neo \citep{clearbot_this_2022}. The Ocean Cleanup company has multiple systems in place to tackle the ocean macroplastics problem. They use satellites to map ocean plastic distribution \citep{biermann_towards_2019}, a global circulation model \citep{torres_partitioning_2018}, data by HYCOM to track currents \citep{hycom_hycom_2023}, Global Forecast System (GFS) for wind tracking \citep{ncei_global_2023} to guide ship fleets with nets to capture macroplastics.\\
\textbf{MARL and Communication Systems:} Communication in Multi-Agent settings is of particular interest as our society and many other distributed systems succeed by collaboration, and therefore this is particularly relevant as most real-world applications include multiple entities with various agencies. \cite{foerster_learning_2016} have been working on sequential prisoner's dilemma \citep{wang_achieving_2019} to investigate collaboration and communication. Social learning has been studied by Jaques et al. with work on emergent social learning \citep{ndousse_emergent_2021}. Further influential work has been an abstraction of Quake 3 Capture the Flag by \cite{jaderberg_human-level_2019}, where agents show collaborative behaviour and develop strategies like following or flag camping. Hide and Seek has been investigated by OpenAi, and results show the emergence of collaborative strategies where agents help each other to build hiding spaces \citep{baker_emergent_2020}. Crucial development on social dilemmas have been made by DeepMind and their work on Melting Pot, a multi-agent environment suite to investigate social dilemmas such as common property resource management \citep{leibo_scalable_2021, agapiou_melting_2022}. Most recent advancements in social behaviour and collaboration has been made on the back of the board game Diplomacy. Both DeepMind \citep{kramar_negotiation_2022} and Meta AI \citep{meta_fundamental_ai_research_diplomacy_team_fair_human-level_2022} have shown great results on agents that were able to negotiate, a crucial aspect for a winning strategy.

\section{Background}
\label{appendix:background}

\textbf{Proximal Policy Optimisation (PPO):} Two main concepts define the Proximity Policy Optimisation (PPO) \citep{schulman_proximal_2017}, a state-of-the-art, on-policy RL algorithm: 1. PPO performs the largest possible but safe gradient ascent learning step by estimating a trust region and 2. Advantage estimates how good an action in a specific state is, compared to the average action. A trust region can be calculated as the quotient of the current policy to be refined $\pi_\theta(a_t|s_t)$ and the previous policy as follows $r_t(\theta) = \frac{\pi_\theta(a_t|s_t)}{\pi_{\theta_k}(a_t|s_t)} = \frac{current\ policy}{old \  policy}$. Advantage is the difference of the Q and the Value Function: $A(s,a) = Q(s,a) - V(s)$, where $s$ is the state and $a$ the action \citep{zychlinski_complete_2019}. The Q function is measures the overall expected reward given state $s$, performing action $a$, and denoted as: $\mathcal{Q}(s,a) = \mathbb{E}\left[ \sum_{n=0}^{N} \gamma^n r_n \right]$. The Value Function, similar to the Q Function, measures overall expected reward, with the difference that the State Value is calculated after the action has been taken and is denoted as: $\mathcal{V}(s) = \mathbb{E}\left[ \sum_{n=0}^{N} \gamma^n r_n \right]$.

\textbf{Graph Neural Network (GNN):} A Graph Neural Network (GNN) is a neural network that operates on graph-structured data, represented by a graph $G = (V, E)$ where $V$ is the set of nodes and $E$ is the set of edges. Each node $v$ in $V$ is associated with a feature vector $x_v$, and the goal of the GNN is to learn a function that maps the graph $G$ and the feature vectors of the nodes to some output $y$. The message passing process in a GNN is typically formalized as a set of iterative updates for the feature vectors of the nodes \citep{gilmer_neural_2017, battaglia_relational_2018}. At each iteration $t$, each node $v$ updates its feature vector $x_v^(t)$ by aggregating information from its neighboring nodes. This process can be mathematically represented as follows: $x_v^{t+1} = f(x_v^{t}, \sum_{u \in N_v} m_{v,u}^{t})$, where $N_v$ is the set of neighbors of node $v$, f is a non-linear function (such as a neural network layer), and $m_{v,u}^{t}$ is a message passed from node $u$ to node $v$ at iteration $t$.

\section{Pseudocode}

PPO-CLIP pseudocode \citep{openai_proximal_2021, schulman_proximal_2017}:

\begin{algorithm}
	\caption{PPO-Clip}
	
	\begin{algorithmic}[1]
		\item Input: initial policy parameters $\theta_0$, initial value function parameters $\phi_0$
		\For {$k=0,1,2,\ldots$}
		    \State Collect set of trajectories $\mathcal{D}_k$ = \{$\tau_i$\} by running policy $\pi_k = \pi(\theta_k)$ in the environment.
		    \State Compute rewards-to-go $\hat{R_t}$.
		    \State Compute advantage estimates, $\hat{A_t}$ (using any method of     advantage estimation) based on the
		    \State current value function $V_{\phi_k}$
			\State Update the policy by maximizing the PPO-Clip objective:
			\State $\theta_{k+1} = arg\underset{\theta}{max} \frac{1}{|\mathcal{D}_k|T} \sum_{\tau \in \mathcal{D}_k} \sum_{t = 0}^{T} \min \left( \frac{\pi_\theta(a_t|s_t)}{\pi_{\theta_k}(a_t|s_t)}A^{\pi_{\theta_k}}(s_t, a_t), g(\epsilon, A^{\pi_{\theta_k}}(s_t, a_t)) \right)$,
			\State typically via stochastic gradient ascent with Adam.
			\State Fit value function by regression on mean-squared error:
			\State $\phi_{k+1} = arg\underset{\phi}{min} \frac{1}{|\mathcal{D}_k|T} \sum_{\tau \in \mathcal{D}_k} \sum_{t = 0}^{T} \left( (V_{\phi}(s_t)-\hat{R_t} \right)$
			\State typically via some gradient descent algorithm.
		\EndFor
	\end{algorithmic} 
\end{algorithm}

Simple Multi-Agent PPO pseudocode:

\begin{algorithm}
	\caption{Multi-Agent PPO} 
	\begin{algorithmic}[1]
		\For {$iteration=1,2,\ldots$}
			\For {$actor=1,2,\ldots,N$}
				\State Run policy $\pi_{\theta_{old}}$ in environment for $T$ time steps
				\State Compute advantage estimates $\hat{A}_{1},\ldots,\hat{A}_{T}$
			\EndFor
			\State Optimize surrogate $L$ wrt. $\theta$, with $K$ epochs and minibatch size $M\leq NT$
			\State $\theta_{old}\leftarrow\theta$
		\EndFor
	\end{algorithmic} 
\end{algorithm}

\newpage
\section{Hyperparameters}
\subsection{Multi Agent Training Hyperparameters}
\begin{verbatim}
environment_parameters:
  # Training: 0, Inference: 1
  scenario_noise_z_shift_factor: 0
  # MA: 0, MAC: 1
  communiation: 0

behaviors:
  PlasticCollector:
    trainer_type: ppo
    hyperparameters:
      batch_size: 512
      buffer_size: 10240
      learning_rate: 0.00001
      beta: 0.01
      epsilon: 0.1
      lambd: 0.95
      num_epoch: 3
      learning_rate_schedule: linear
    network_settings:
      normalize: false
      hidden_units: 512
      num_layers: 2
      vis_encode_type: simple
    reward_signals:
      extrinsic:
        gamma: 0.99
        strength: 1.0
    keep_checkpoints: 5
    max_steps: 20000000
    time_horizon: 128
    summary_freq: 10000
\end{verbatim}

\newpage
\subsection{Hyperparameter Description}
\label{appendix:hyper}
\begin{table}[ht]
\centering
    \begin{tabular}{p{0.3\textwidth}p{0.3\textwidth}p{0.3\textwidth}}
    \toprule
    %\multicolumn{2}{c}{Part}                   \\
    %\cmidrule(r){1-2}
    Hyperparameter & Typical Range & Description\\
    \midrule
    Gamma & $0.8-0.995$ & discount factor for future rewards\\
    Lambda & $0.9-0.95$ & used when calculating the Generalized Advantage Estimate (GAE)\\
    Buffer Size & $2048-409600$ & how many experiences should be collected before updating the model\\
    Batch Size & $512-5120$ (continuous), $32-512$ (discrete) & number of experiences used for one iteration of a gradient descent update.\\\
    Number of Epochs & $3-10$ & number of passes through the experience buffer during gradient descent\\
    Learning Rate & $1e-5-1e-3$ & strength of each gradient descent update step\\
    Time Horizon & $32-2048$ & number of steps of experience to collect per-agent before adding it to the experience buffer\\
    Max Steps & $5e5-1e7$ & number of steps of the simulation (multiplied by frame-skip) during the training process\\
    Beta & $1e-4-1e-2$ & strength of the entropy regularization, which makes the policy "more random"\\
    Epsilon & $0.1-0.3$ & acceptable threshold of divergence between the old and new policies during gradient descent updating\\
    Normalize & $true/false$ & weather normalization is applied to the vector observation inputs\\
    Number of Layers & $1-3$ & number of hidden layers present after the observation input\\
    Hidden Units & $32-512$ & number of units in each fully connected layer of the neural network\\
    \midrule
    Intrinsic Curiosity Module\\
    \midrule
    Curiosity Encoding Size & $64-256$ & size of hidden layer used to encode the observations within the intrinsic curiosity module\\
    Curiosity Strength & $0.1-0.001$ & magnitude of the intrinsic reward generated by the intrinsic curiosity module\\
    \bottomrule
    \end{tabular}
\caption{Hyperparameters Description: \url{https://github.com/Unity-Technologies/ml-agents/blob/main/docs/Training-Configuration-File.md}}
\label{sample-table}
\end{table}

\newpage
\section{Ocean Plastic Collector Environment Scenario Samples 0-19}
\begin{figure}[ht]
\centering
    \includegraphics[width=0.95\textwidth]{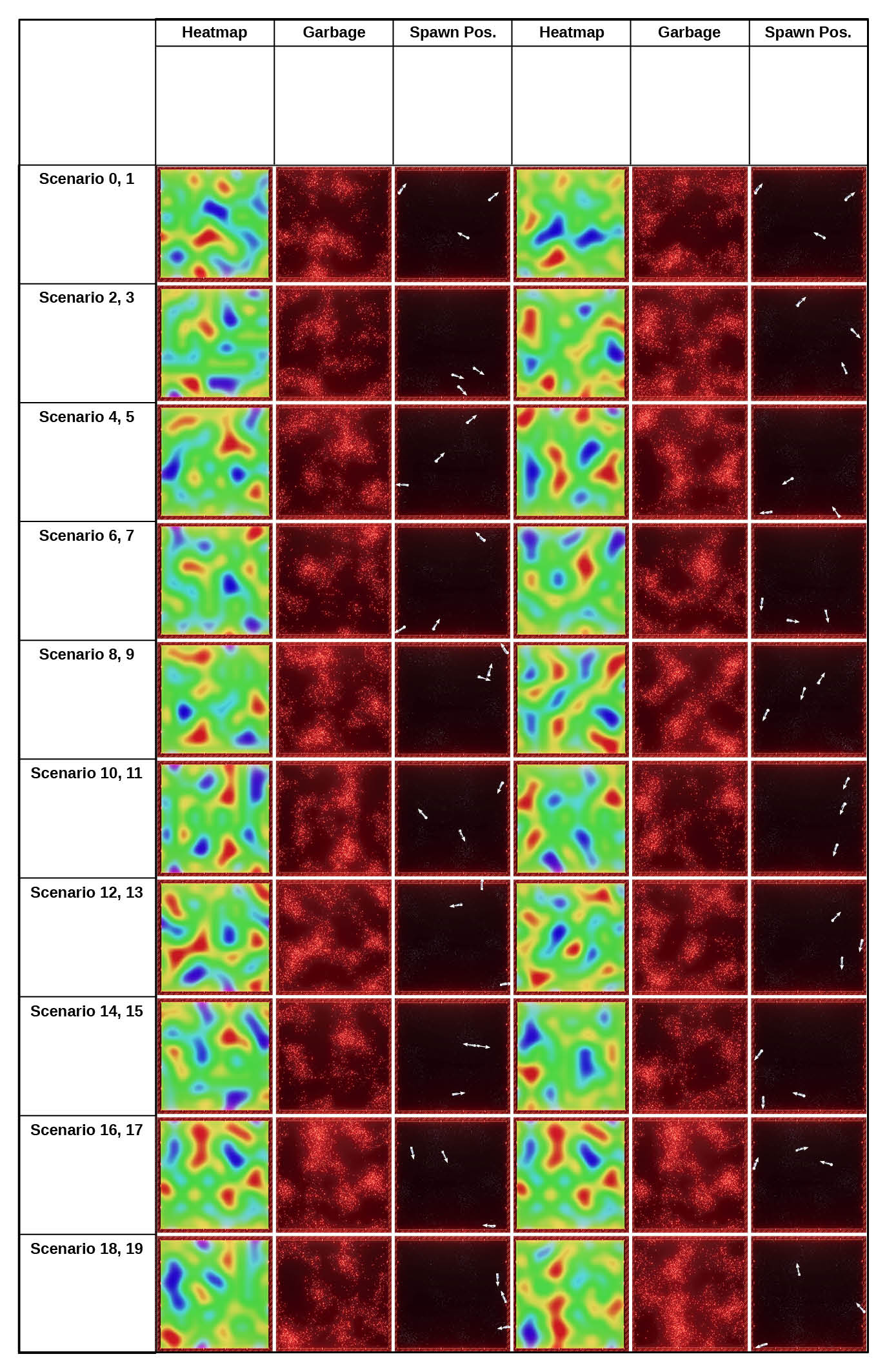}
\caption{Scenario Samples 0-19; Heatmap, Garbage Distribution, Random Spawn Position}
\label{appendix:env-samples}
\end{figure}

\newpage

\section{Ocean Plastic Collector Environment: Main Features}
\label{appendix:scenario-samples}

\begin{figure}[ht]
\centering
    \begin{subfigure}{0.3\textwidth}
    \includegraphics[width=\textwidth]{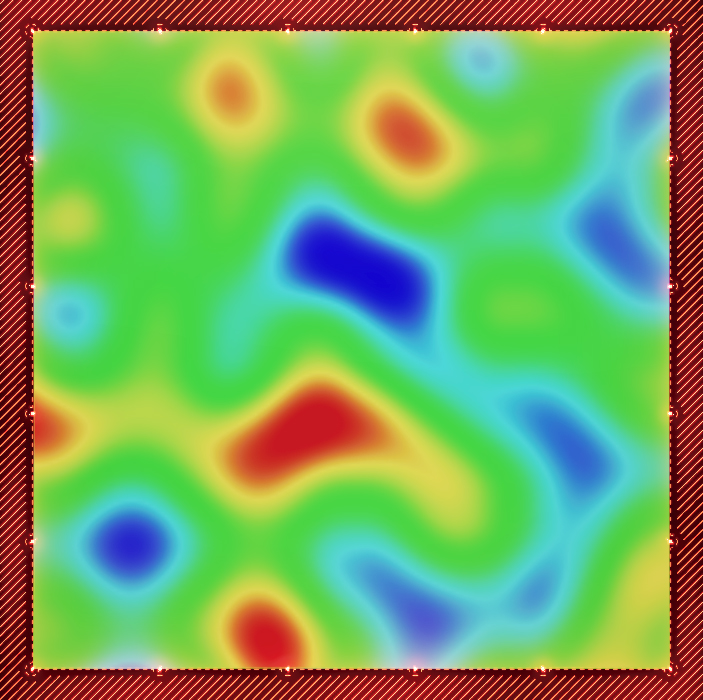}
    \caption{Garbage Distribution Heatmap}
    \end{subfigure}
    \begin{subfigure}{0.3\textwidth}
    \includegraphics[width=\textwidth]{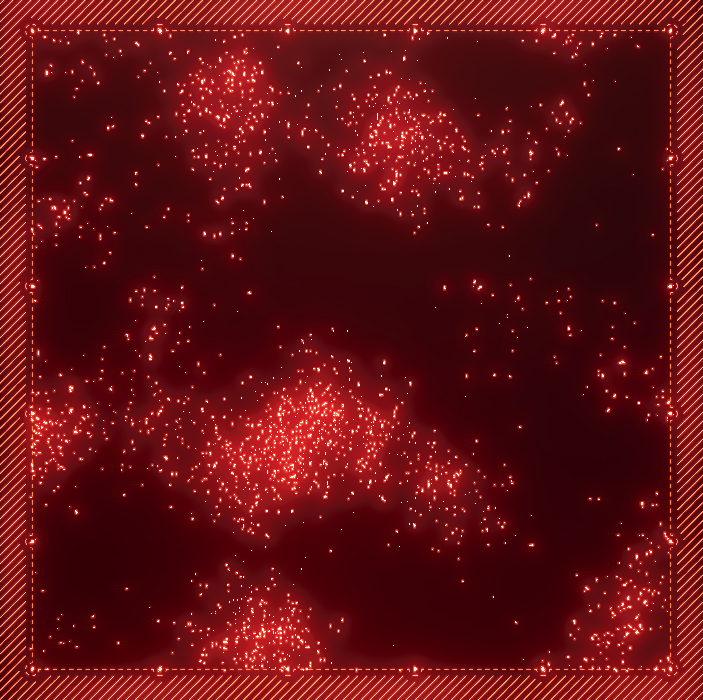}
    \caption{Distributed Garbage}
    \end{subfigure}
    \begin{subfigure}{0.3\textwidth}
    \includegraphics[width=\textwidth]{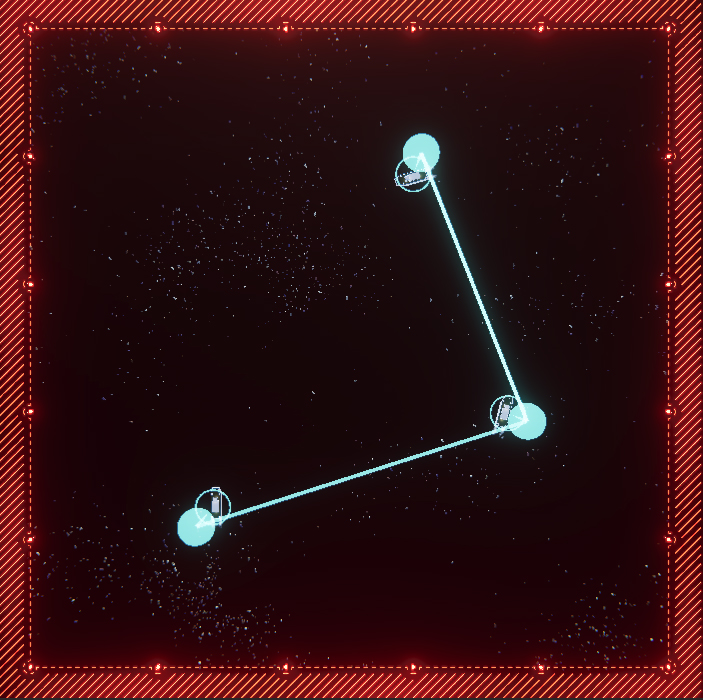}
    \caption{Agent Com. Network}
    \end{subfigure}
\caption{Plastic Collector Environment: Main Features}
\label{appendix:env-features}
\end{figure}

\begin{figure}[ht]
\centering
    \includegraphics[width=0.85\textwidth]{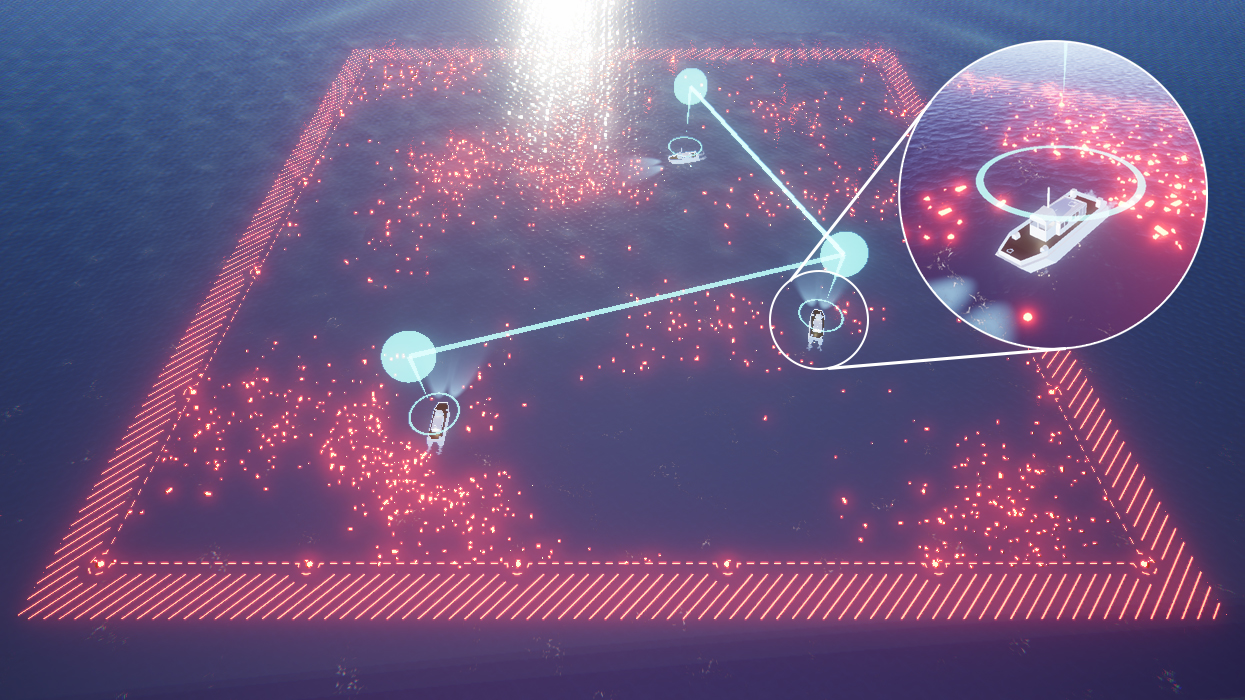}
\caption{Plastic Collector Environment: Garbage and Vessel Communication Network}
\label{appendix:training-zoom-in}
\end{figure}

\begin{figure}[ht]
\centering
    \includegraphics[width=0.85\textwidth]{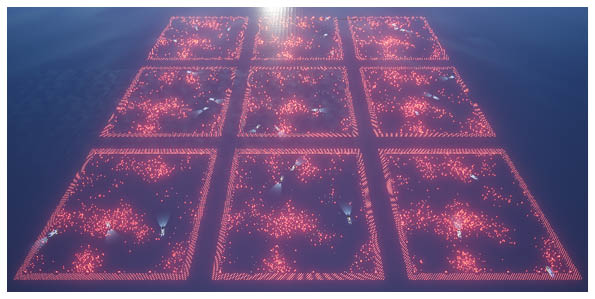}
\caption{Plastic Collector Environment: Training School}
\label{appendix:training-field}
\end{figure}

\newpage
\section{Multi-Agent Process Diagram}

\begin{figure}[ht]
\centering
    \includegraphics[width=\textwidth]{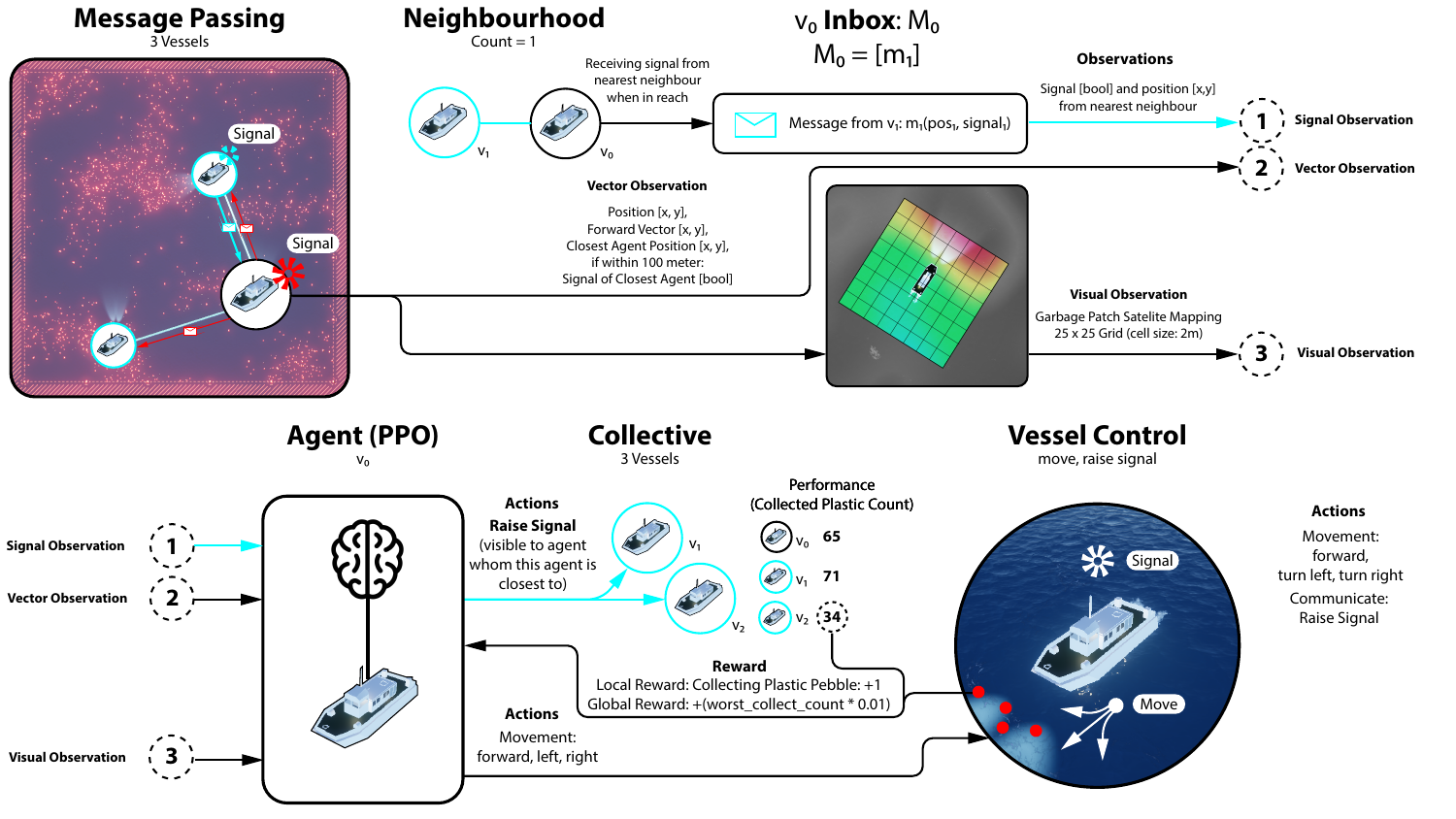}
\caption{Multi-Agent Process Diagram}
\label{appendix:process}
\end{figure}

\section{Training Graphs}

\begin{figure}[H]
\centering
    \begin{subfigure}{0.45\textwidth}
    \includegraphics[width=\textwidth]{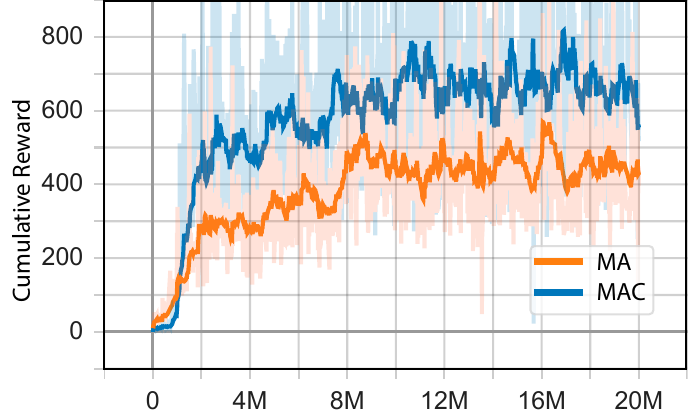}
    \caption{Cumulative Reward}
    \end{subfigure}
    \begin{subfigure}{0.45\textwidth}
    \includegraphics[width=\textwidth]{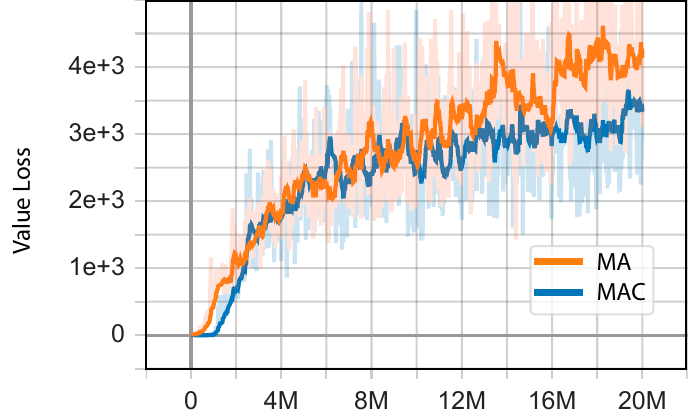}
    \caption{Value Loss}
    \end{subfigure}
    \begin{subfigure}{0.45\textwidth}
    \includegraphics[width=\textwidth]{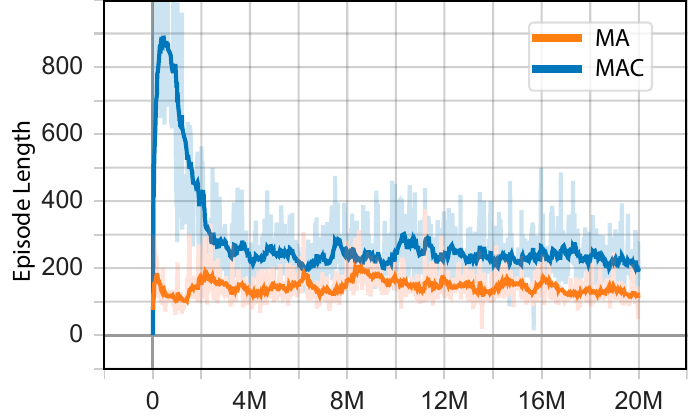}
    \caption{Episode Length}
    \end{subfigure}
    \begin{subfigure}{0.45\textwidth}
    \includegraphics[width=\textwidth]{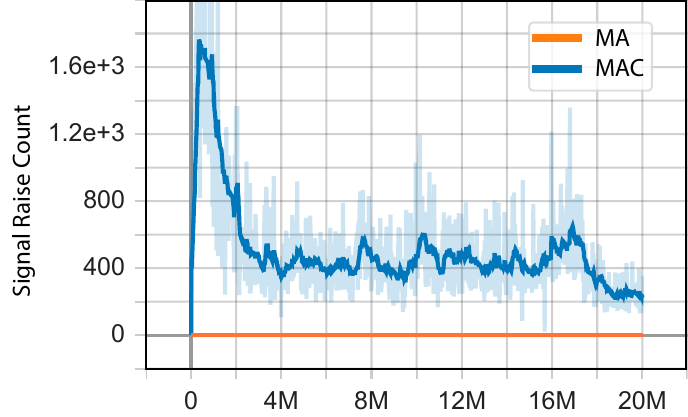}
    \caption{Signal Raise Count}
    \end{subfigure}
\caption{Train graphs: Cumulative Reward, Value Loss, Episode Length, Signal Raise Count}
\label{appendix:train-data-1}
\end{figure}

\newpage

\begin{figure}[H]
\centering
    \begin{subfigure}{0.45\textwidth}
    \includegraphics[width=\textwidth]{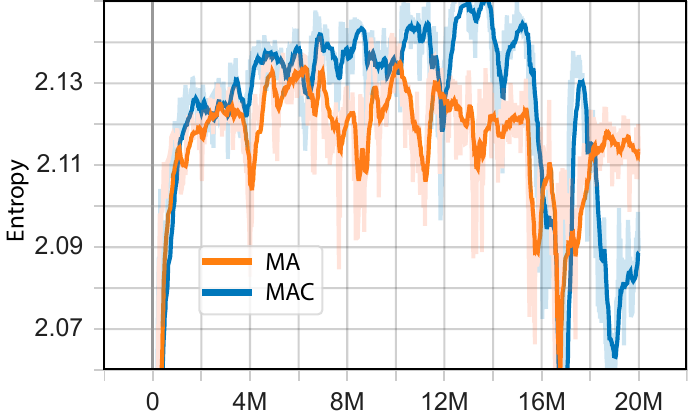}
    \caption{Entropy}
    \end{subfigure}
    \begin{subfigure}{0.45\textwidth}
    \includegraphics[width=\textwidth]{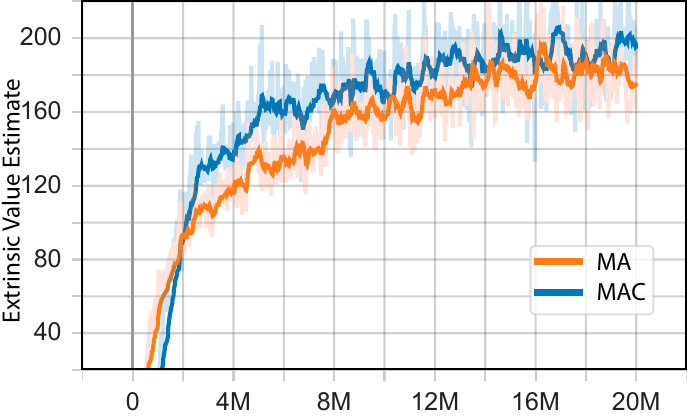}
    \caption{Extrinsic Value Estimate}
    \end{subfigure}
\caption{Train Data Plot: Entropy, Extrinsic Value Estimate}
\label{appendix:train-data-2}
\end{figure}

% TEST DATA
\section{Testing Graphs}

\begin{figure}[ht]
\centering
    \begin{subfigure}{0.45\textwidth}
    \includegraphics[width=\textwidth]{source/inference_results/test_results_Environment_Cumulative_Reward.png}
    \caption{Cumulative Reward}
    \end{subfigure}
    \begin{subfigure}{0.45\textwidth}
    \includegraphics[width=\textwidth]{source/inference_results/test_results_Environment_Episode_Length.png}
    \caption{Episode Length}
    \end{subfigure}
    \begin{subfigure}{0.45\textwidth}
    \includegraphics[width=\textwidth]{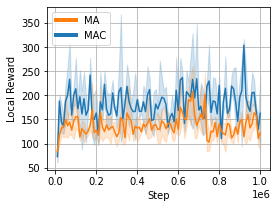}
    \caption{Local Reward: Per Garbage Pebble Collected}
    \end{subfigure}
    \begin{subfigure}{0.45\textwidth}
    \includegraphics[width=\textwidth]{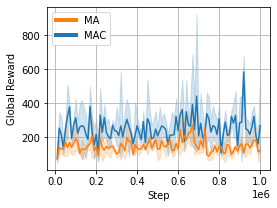}
    \caption{Global Reward}
    \end{subfigure}
\caption{Testing graphs: Cumulative Reward, Episode Length, Local and Global Reward}
\label{appendix:test-data-basics}
\end{figure}

\newpage

\begin{figure}[ht]
\centering
    \begin{subfigure}{0.45\textwidth}
    \includegraphics[width=\textwidth]{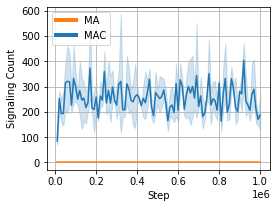}
    \caption{Signal Count per Episode}
    \end{subfigure}
    \begin{subfigure}{0.45\textwidth}
    \includegraphics[width=\textwidth]{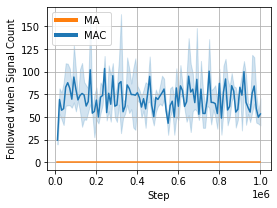}
    \caption{Followed Count when Signal: 1}
    \end{subfigure}
    \begin{subfigure}{0.45\textwidth}
    \includegraphics[width=\textwidth]{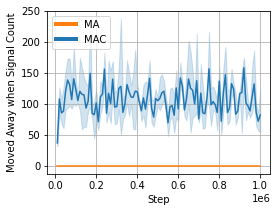}
    \caption{Moved Away Count when Signal: 1}
    \end{subfigure}
    \begin{subfigure}{0.45\textwidth}
    \includegraphics[width=\textwidth]{source/inference_results/test_results_action_on_signal.png}
    \caption{Action on Signal: 1: Follow, Move Away }
    \end{subfigure}
    \begin{subfigure}{0.45\textwidth}
    \includegraphics[width=\textwidth]{source/inference_results/test_results_Nearby_Garbage_Count.png}
    \caption{Neraby Garbage Count when Signal: 1 and 0}
    \end{subfigure}
\caption{Testing graphs: Signal}
\label{appendix:test-data-signal}
\end{figure}

\newpage

\section{Vessel Path Scenario 1-3: Garbage Nearby Count \& Signal Communication}

\begin{figure}[ht]
\centering
    \begin{subfigure}{0.37\textwidth}
    \includegraphics[width=\textwidth]{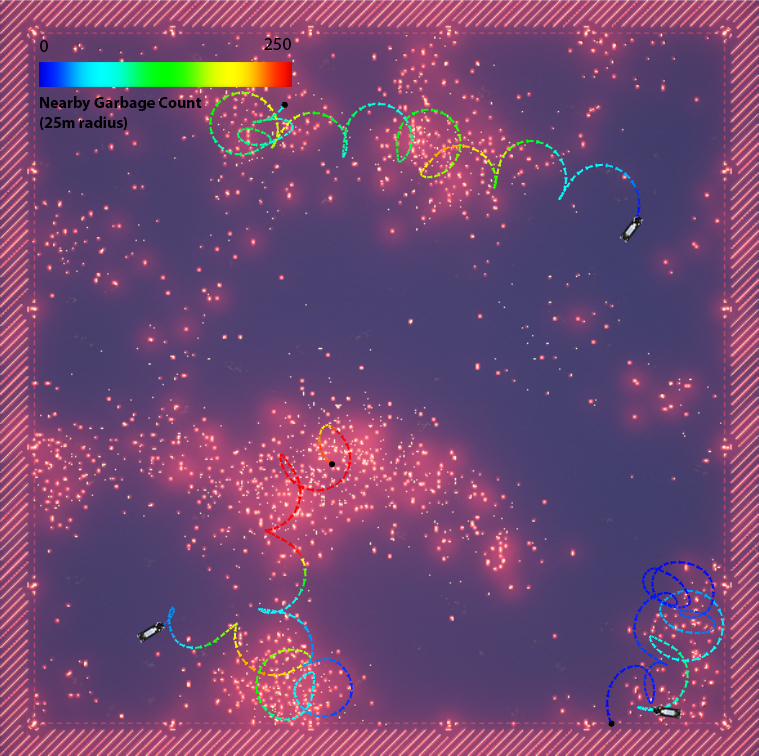}
    \caption{Garbage Count within 25m radius}
    \end{subfigure}
    \begin{subfigure}{0.37\textwidth}
    \includegraphics[width=\textwidth]{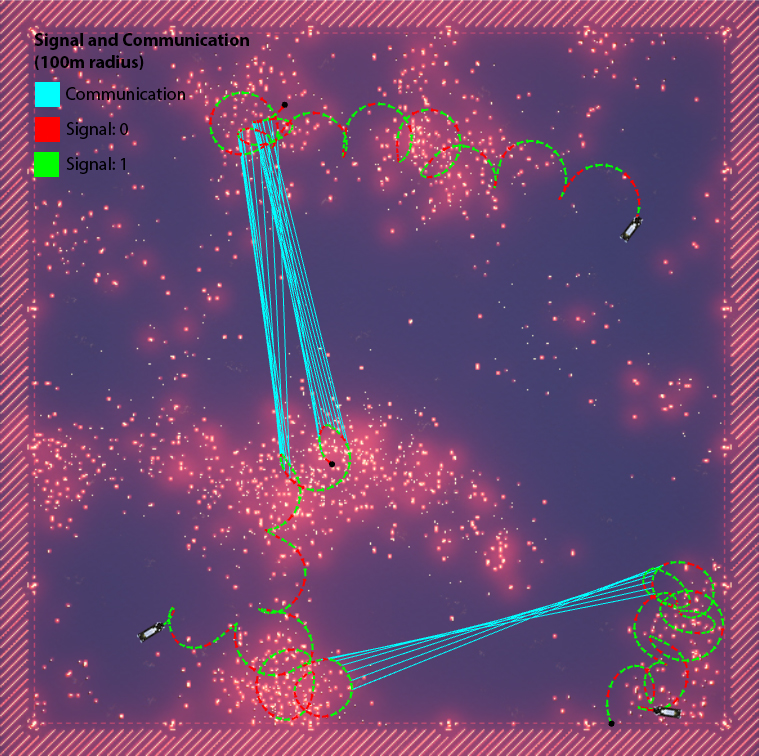}
    \caption{Signal 0, 1 and Communication}
    \end{subfigure}
\caption{Scenario 1: Vessel Path}
\label{appendix:path-1}
\end{figure}
\begin{figure}[ht]
\centering
    \begin{subfigure}{0.37\textwidth}
    \includegraphics[width=\textwidth]{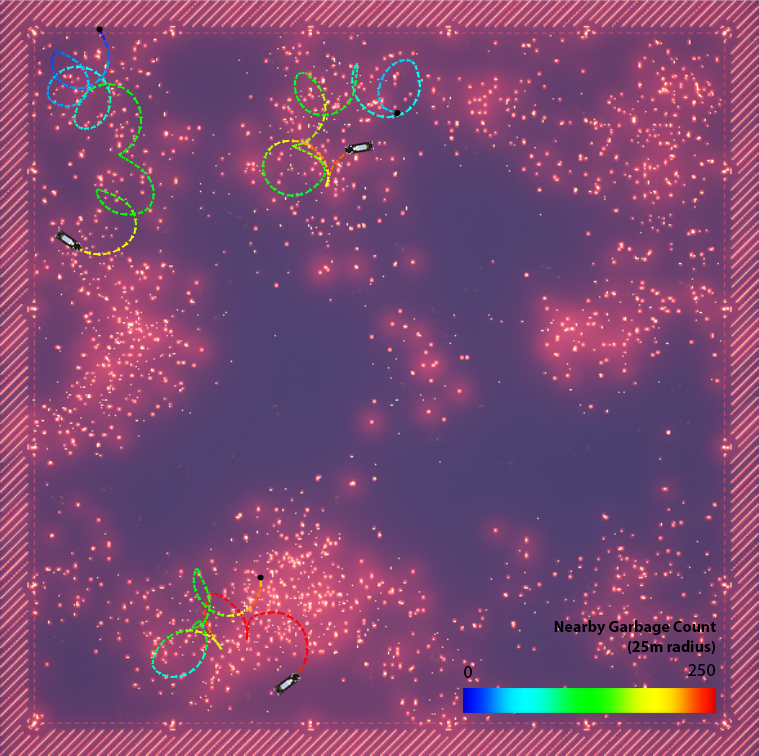}
    \caption{Garbage Count within 25m radius}
    \end{subfigure}
    \begin{subfigure}{0.37\textwidth}
    \includegraphics[width=\textwidth]{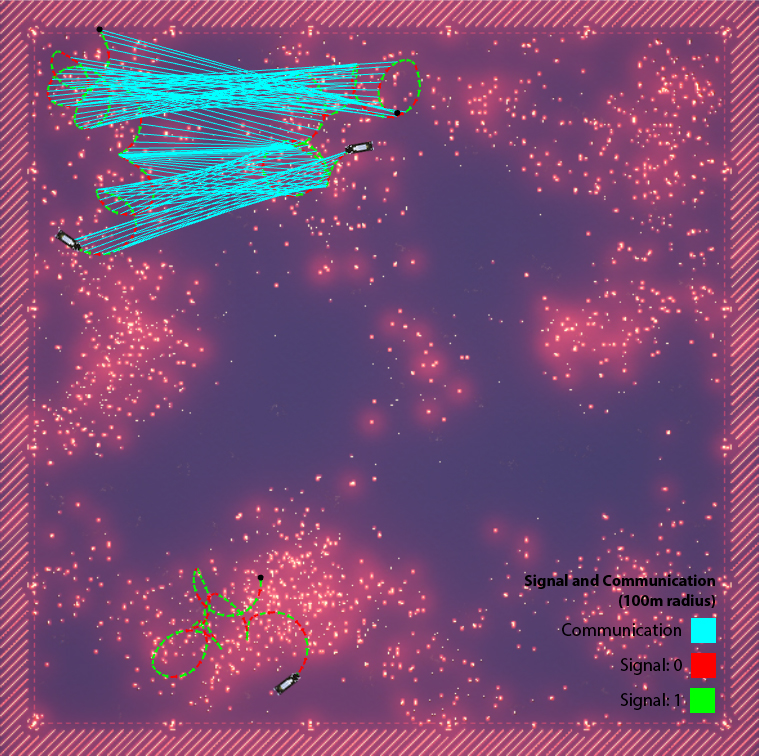}
    \caption{Signal 0, 1 and Communication}
    \end{subfigure}
\caption{Scenario 2: Vessel Path}
\label{appendix:path-2}
\end{figure}
\begin{figure}[ht]
\centering
    \begin{subfigure}{0.37\textwidth}
    \includegraphics[width=\textwidth]{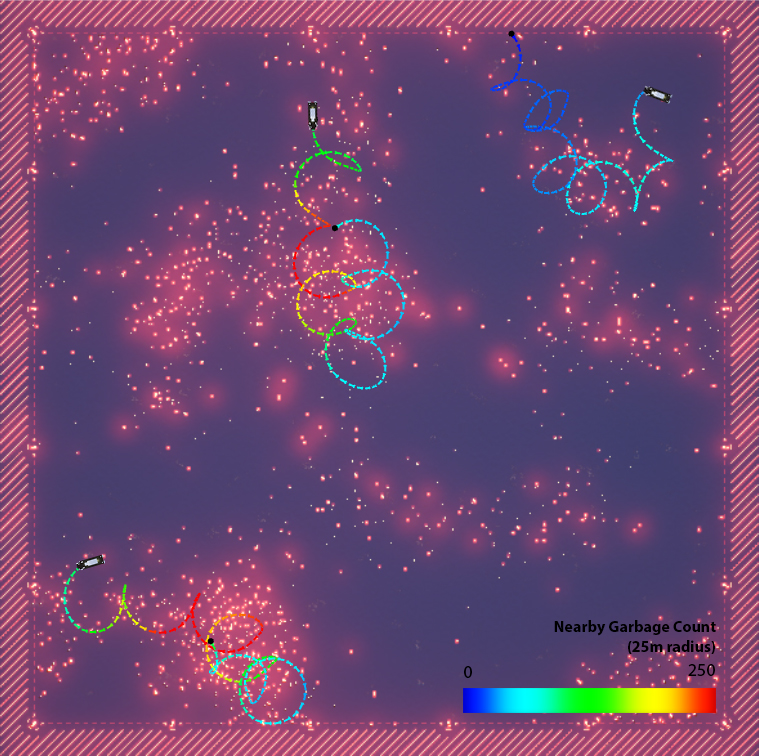}
    \caption{Garbage Count within 25m radius}
    \end{subfigure}
    \begin{subfigure}{0.37\textwidth}
    \includegraphics[width=\textwidth]{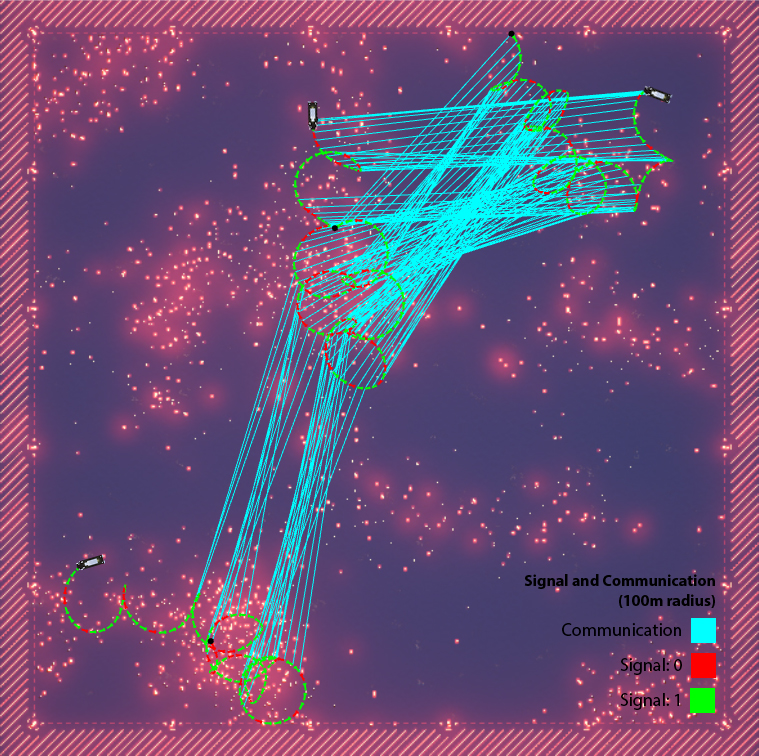}
    \caption{Signal 0, 1 and Communication}
    \end{subfigure}
\caption{Scenario 3: Vessel Path}
\label{appendix:path-3}
\end{figure}

\end{document}